\documentclass[11pt]{article}

\usepackage{jair}
\usepackage{theapa}
\usepackage{graphics}
\usepackage{epsfig}

\jairheading{26}{2006}{417-451}{11/05}{08/06}
\ShortHeadings{Multiple-Goal Heuristic Search}
{Davidov \& Markovitch}
\firstpageno{417}

\newcommand{\TABLINE}[1]{\hspace*{#1}\=\hspace{#1}\=\hspace{#1}\=\hspace{#1}\=\hspace{#1}\=\hspace{#1}\=\hspace{#1}\=\hspace{#1}\=\hspace{#1}\=\hspace{#1}\=\hspace{#1}\=
\kill}

\newcommand{\PM}[3]{$\scriptsize$(#1#2#3)$\normalsize $}

\def\twofig#1#2{%
\epsfxsize=0.48\textwidth \noindent \epsffile{#1}\hfill
\epsfxsize=0.48\textwidth
\epsffile{#2}\\
\makebox[0.48\textwidth]{(a)}\hfill\makebox[0.48\textwidth]{(b)}%
}

\newtheorem{Definition}{Definition}

\begin{document}

\title{
Multiple-Goal Heuristic Search
}

\author{\name Dmitry Davidov \email dmitry@cs.technion.ac.il\\
\name Shaul Markovitch \email shaulm@cs.technion.ac.il\\
\addr Computer Science Department\\
Technion, Haifa 32000, Israel}

\maketitle

\begin{abstract}

This paper presents a new framework for anytime heuristic search
where the task is to achieve as many goals as possible within the
allocated resources. We show the inadequacy of traditional
distance-estimation heuristics for tasks of this type and present
alternative heuristics that are more appropriate for multiple-goal
search.  In particular, we introduce the marginal-utility
heuristic, which estimates the cost and the benefit of exploring a
subtree below a search node. We developed two methods for online
learning of the marginal-utility heuristic. One is based on local
similarity of the partial marginal utility of sibling nodes, and
the other generalizes marginal-utility over the state feature
space. We apply our adaptive and non-adaptive multiple-goal search
algorithms to several problems, including focused crawling, and
show their superiority over existing methods.

\end{abstract}

\section{Introduction}
 Internet search engines build their indices using brute-force
crawlers that attempt to scan large portions of the Web. Due to
the size of the Web, these crawlers require several weeks to
complete one scan, even when using very high computational power
and bandwidth~\cite{brin98anatomy,douglis97rate}, and they still
leave a large part of the Web uncovered~\cite{lawrence98searching,Bharat:1998}. Many times, however, it
is necessary to retrieve only a small portion of Web pages dealing
with a specific topic or satisfying various user criteria. Using
brute-force crawlers for such a task would require enormous
resources, most of which would be wasted on irrelevant pages. Is
it possible to design a \emph{focused} crawler that would scan
only relevant parts of the Web and retrieve the desired pages
using far fewer resources than the exhaustive crawlers?

Since the Web can be viewed as a large graph~\cite{cooper02crawling,kumar00web,fp+pr+eu:pagerank_web_structure},
where pages are nodes and links are arcs, we may look for a
solution to the above problem in the field of heuristic
graph-search algorithms.  A quick analysis, however, reveals that
the problem definition assumed by the designers of heuristic
search algorithms is inappropriate for focused crawling, which
uses an entirely different setup. The crucial difference between
heuristic search and focused crawling is the success criterion. In
both setups there is a \emph{set} of goal states. In the heuristic
search setup, however, the search is completed as soon as a single
goal state is found, while in the focused crawling setup,  the
search continues to reach as many goal states as possible within
the given resources.

\begin{figure}[!htb]
\centering \scalebox{0.4}{\includegraphics{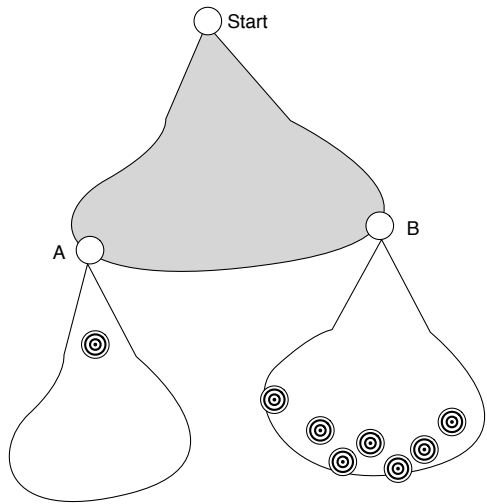}}
\caption{Using a distance-estimation heuristic for a multiple-goal
search problem}\label{heuristic-example}
\end{figure}

Changing the success criterion of existing search algorithms is
not enough. Most informed search algorithms are based on a
heuristic function that estimates the distance from a node to the
nearest goal node. Such heuristics are usually not appropriate for
multiple-goal search. Consider the search graph described in
Figure \ref{heuristic-example}.
The grey area is the expanded graph. Assume that we evaluate nodes
$A$ and $B$ using a distance-based heuristic.  Node $A$ has a
better heuristic value and will therefore be selected. This is
indeed the right decision for traditional search where the task is
to find \emph{one} goal. For multiple-goal search, however, $B$
looks like a much more promising direction since it leads to an
area with a high density of goal nodes.

For many problems in a wide variety of domains
 where we are interested in finding a set of goals rather
 than a single goal. In genetic
engineering, for example,  we want to find multiple possible
alignments of several DNA sequences~\cite{yoshizumi00with,Korf:2000}. In chemistry we may want to
find multiple substructures of a complex molecule. In robotics, we
may want to plan paths for multiple robots to access multiple
objects. In some of these cases, a possible solution would be to
invoke single-goal search multiple times. Such an approach,
however, is likely to be wasteful, and for resource-bounded
computation, we may wish to exploit the multiple-goal\footnote{
Note that the term ``multiple-goal'' is also used in the planning
domain. There, however, the task is to satisfy a set of dependent
goals with possible order constraints. } nature of the problem to
make the search more efficient.

The specific multiple-goal task of focused crawling has received
much attention~\cite{chakrabarti99focused,cho00evolution,cho98efficient,diligenti00focused,rennie99using}
because of the popularity of the Web domain. Most of these works,
however, focused on Web-specific techniques tailored for the
particular problem of crawling.

The goal of the research described in this paper is to establish a
new domain-independent framework for multiple-goal search problems
and develop anytime heuristic algorithms for solving them
efficiently. Our framework focuses mainly on problem domains where
we are looking for the goal states and the paths to the goals
either are irrelevant, or their cost is of no concern (except for
its effect on the search cost).

We start with a formal definition of the multiple-goal search
problem.  We then describe versions of existing heuristic search
algorithms, modified for the multiple-goal framework. The main
differences between single-goal and multiple-goal search are the
heuristic functions used.  We describe a new set of heuristics
that are better suited for multiple-goal search.  In particular,
we introduce the marginal-utility heuristic, which considers the
expected costs as well as the expected benefits associated with
each search direction. It is difficult to specify such heuristics
 explicitly.  We therefore present adaptive methods that allow online
learning of them.  Finally we describe an extensive empirical
study of our algorithms in various domains, including the focused
crawling problem.

The contributions of this paper are fourfold:
\begin{enumerate}
\item We identify and define the framework
of multiple-goal heuristic search.  This is a novel framework for heuristic
search.
\item We define a set of search algorithms and heuristics that are appropriate for multiple-goal
search problems.
\item We define a utility-based heuristic and present a method for
automatic acquisition of it via online
learning.
\item We provide extensive empirical study of the presented methods
in various domains and show their superiority over existing general
algorithms.
\end{enumerate}


\section{The Multiple-Goal Search Problem}
   \label{sec-problem}
 Let $\langle S, E \rangle$ be a potentially
infinite state graph with finite degree, where $S$ is a set
of states and $E \subseteq S \times S$ is a set of edges.
 The \emph{single-goal
search problem} can be defined as follows:
\begin{enumerate}
\item \emph{Input:}
    \begin{itemize}
        \item{A set of initial states $S_i \subseteq S$}
       \item A successor function $\mbox{Succ} : S \rightarrow 2^S$
 such that $\mbox{Succ}(s) = \{ s' \: | \: \langle s , s' \rangle \in E \}$.
(Sometimes Succ is given implicitly be a finite set of operators $O$.)
        \item A goal predicate $G: S \rightarrow \{0,1\}$.
We denote $S_g = \{s \in S \: | \:G(s)\}$.
Sometimes $S_g$ is given explicitly.
      \end{itemize}
\item \emph{Search objective:}
   Find a goal state $g \in S_g$ such that there is a directed path in
$\langle S, E \rangle$ from a state in $S_i$ to $g$.
  Sometimes we are also interested in the path itself.
\item \emph{Performance evaluation:}  Although the performance
evaluation criterion is not commonly considered to be an integral
part of the problem definition, we do consider it as such. This is
because it determines the class of algorithms to be considered.
The most common criteria for evaluating a search are the solution
quality, usually measured by the solution path cost, and search
efficiency, mostly evaluated by the resources consumed during the
search.
\end{enumerate}
 Although the \emph{multiple-goal search problem} bears some similarity to the single-goal search problem, it differs in several
 ways:
\begin{enumerate}
\item \emph{Input:} Includes an additional resource limit $R$. For
simplicity of presentation we assume that $R$ is given to us as a
number of generated nodes\footnote{ Our framework is therefore
applicable to any resource that is proportional to the number of
generated nodes.  That can be CPU time, internet bandwidth, energy
consumption by robots, etc.  For best-first search algorithms, the
number also corresponds to the memory consumption.} . Later we
will discuss this assumption. \item \emph{Search objective:} Find
a \emph{set} of goal states $S^R_g \subseteq S_g$ which satisfies:
\begin{itemize}
\item  For each $s \in S^R_g$, there is a directed path in
 $\langle S, E \rangle$ from some $s_i \in S_i$ to $s$.
\item The search resources consumed should not exceed $R$.
\end{itemize}
  Sometimes we are also interested in the set of corresponding paths.
\item \emph{Performance evaluation:}  $\left | S^R_g \right |$.
Obviously, higher values are considered better.
\end{enumerate}

While it looks as if we mix here reasoning with meta reasoning by
inserting resource limit as part of problem input, many problems
are much more naturally defined with such a resource limitation.
Consider, for example, the minimax algorithm, where the maximal
depth of search (which determines the resources consumed) is given
as part of the algorithm's input. The above formulation, where the
resource limit is given as an input, falls under the scope of
problems solved by \emph{anytime algorithms}
 \cite{Boddy:1994:DSP,Hovitz:1990:CAB,Zilberstein:1996:UAA} and
more specifically by \emph{contract
algorithms}~\cite{Russell:1991:CRTS,zilberstein99realtime}.

Sometimes the resource limit is not known in advance. In such a
setup, the search algorithm can be interrupted at any time and
required to return the current set of collected goals.  This type
of problem is solved by \emph{interruptible anytime
algorithms}~\cite{hansen96monitoring,Russell:1991:CRTS}. An
alternative formalization is to require the algorithm to find a
specified number of goals and to evaluate its performance by the
resources consumed during the search.

\section{Multiple-Goal Heuristic Search Algorithms}
\label{searchalgs}
 Assume that $h_{mg}:S\rightarrow \Re$ is a heuristic function
that estimates the merit of states with respect to the
objective and performance evaluation criterion
of a multiple-goal search problem as defined in the previous
section.

Our objective is to develop search algorithms that can exploit
such heuristics in a similar manner to the heuristic search
algorithms developed for single-goal search problems.
We start by describing the multiple-goal version of greedy
 best-first search\footnote{We follow~\citeA[page 95]{Russell-Norvig-2nd}
 who use this term to describe a search algorithm that always expands the
 node estimated to be closest to a goal.}.

There are two main differences between the existing single-goal
best-first search algorithm and our newly defined
multiple-goal version:
 \begin{enumerate}
 \item While single-goal best-first search stops as soon as it
 encounters a goal state, the multi-goal version collects
 the goal and continues until the allocated resources
 are exhausted.
 \item While single-goal best-first search typically uses
 a heuristic function that tries to approximate the distance
 to the nearest goal, the multiple-goal version should use
 a different type of heuristic that is more appropriate
 for the multiple-goal task.
 \end{enumerate}

Most of the other heuristic search algorithms can also be modified
to find multiple goals. The $A^*_\epsilon$ algorithm
by~\citeA{Pearl:1982} can be converted to handle multiple goal
search by
 collecting each
goal it finds in its \emph{focal} list. These are goals with
$\epsilon-$optimal paths. A multiple-goal heuristic can be used to
select a node from the \emph{focal} list for expansion. The
algorithm stops, as before, when the allocated resources are
exhausted.  In addition, we can stop the algorithm when all the
nodes in \emph{focal} satisfy $f(n) > (1 + \epsilon)g_{min}$,
where $g_{min}$ is the minimal $g$ value among the collected
goals.

Multiple-goal hill-climbing uses a multiple-goal heuristic to
choose the best direction.  We modify this algorithm to allow the
search to continue after a goal is found. One possible method for
continuing the search is to perform a random walk from the found
goal.

Multiple-goal backtracking works similarly to the single goal
version.  However, when a goal is encountered, the algorithm
simulates failure and therefore continues. A multiple goal
heuristic can be used for ordering the operators in each node. For
constraint satisfaction, that means ordering the values associated
with a variable.

\section{Heuristics For Multiple-Goal Problems}
\label{sec-heuristics}
 In the introduction, we illustrated the problem of using a traditional
distance-estimation heuristic for multiple-goal search.  One of
the main problems with using such a distance-estimation heuristic
is that it does not take into account goal density but only the
distance to the nearest goal.  This can lead the search into a
relatively futile branch, such as the left branch in Figure
\ref{heuristic-example}, rather than the much more fruitful right
branch. In this section we consider several alternative heuristic
functions that are more appropriate for multiple-goal search.
\subsection{The Perfect Heuristic}
\label{sec-perfect}
 Before we describe and analyze heuristic functions for multiple-goal search, we would like to
consider the function that we are trying to approximate.  Assume, for
example, that we perform multiple-goal greedy best-first search
and that we have perfect knowledge of the search graph. What node
would we like our heuristic to select next?
Assume that the given resource limit allows us to expand additional $M$
nodes, and look at all the search forests\footnote{
While the search space is a graph, the search algorithm expands
a forest under the currently open nodes.
}
 of size $M$ rooted at the current
list of open nodes.  A perfect heuristic will select a node belonging to
the forest with the largest number of goals.

\begin{Definition}
 Let $S_{\mathit{open}} \subseteq S$ be the set of currently open states.
 Let $R$ be the resource limit and
 $R_c$ be the resources consumed so far.  Let $S_{gf} \subseteq S_g$
 be the set of goals found so far.
 Let $F$ be a set of all possible forests of size $R-R_c$ starting
 from roots in $S_{open}$.  A forest $f \in F$ is  \emph{optimal}
 if and only if
 \[
 \forall f'\in F,
\left|\left(S_g \setminus S_{gf}\right) \cap f'\right|\leq
\left|\left(S_g \setminus S_{gf}\right)  \cap f\right|
 .\]

A state $s \in S_{open}$ is \emph{optimal}, denoted as $OPT(s)$,
if and only if there exists an optimal forest $f$ such that $s \in
f$.
\end{Definition}

\begin{Definition}
A heuristic function $h$ is \emph{perfect} with respect to a
multiple-goal search problem if for every possible search stage
defined by $S_{open}$,

\[
\forall s_1,s_2 \in S_{open} \left [ \mathrm{OPT}(s_1) \wedge \neg \mathrm{OPT}(s_2)
\Longrightarrow h(s_1) < h(s_2) \right ] .\]
\end{Definition}

Thus a perfect heuristic never selects for expansion a state that
is not in an optimal forest. Using such a heuristic in
multiple-goal best-first search will make the search optimal.
Note that the optimality is with respect to search resources
and not with respect to the cost of paths leading to the goal
states.

Obviously, the above definition does not lead to a practical
multiple-goal heuristic. Even for simple problems, and even when
we have perfect knowledge about the graph, calculating such a
heuristic is very hard. This is because the number of possible
forests is exponential in the resource limit and in the number of
nodes of the open list.

\subsection{Sum Heuristics}
\label{sec-sum}
 Many search algorithms that look for a single goal
state use a heuristic function that estimates the cost of the
cheapest path to a goal state. Optimizing algorithms such as
$A^{*}$ require admissible heuristics (that underestimate the real
distance to the goal) while \emph{satisficing} search algorithms,
such as greedy best-first, can use non-admissible heuristics as
well. Distance-estimation heuristics were therefore developed for
many domains. In addition, several researchers have developed
automatic methods for inferring admissible heuristics by
relaxation~\cite{Held:1970,Mostow:1989,Prieditis:1993} and by
pattern databases~\cite{Culberson:1998,Gasser:1995,korf:2002}.

Figure \ref{heuristic-example} illustrates that a straightforward
use of distance heuristics for multiple-goal search is not
appropriate. Here we define a method for utilizing (admissible or
non-admissible) distance heuristics for multiple-goal search.
 Assume that the set of goal states, $S_g$, is given explicitly,
and that we are given a common distance-estimation heuristic
$h_{dist}(s_1,s_2)$ that estimates the graph distance between two
given states\footnote{Assume that the Euclidean distance in the figure
reflects the heuristic distance.}. The \emph{sum-of-distances} heuristic, denoted as
$h_{sum}$, estimates the sum of distances to the goal set:
\begin{equation}
h_{sum}(s)=\sum_{g\in S_g}{h_{dist}(s,g)}.
\label{EQSUM}
\end{equation}
Minimizing this heuristic will bias the search towards larger
groups of goals, thus selecting node $B$ in the example
of Figure \ref{heuristic-example}.
 This is indeed a better
decision, provided that enough resources are left for reaching the
goals in the subgraph below $B$.
\subsection{Progress Heuristics}
\label{sec-progr}
 One problem with the \emph{sum} heuristic is its
tendency to try to progress towards all of the goals
simultaneously. Hence, if there are groups of goals scattered
around the search front, all the states around the front will have
similar heuristic values. Each step reduces some of the distances
and increases others, leading to a more-or-less constant sum. In
such constant-sum regions, an algorithm which relies only on the
sum heuristic will have no information about which node to choose
for expansion.
\begin{figure}[htb]
 \centering
\scalebox{0.8}{\includegraphics{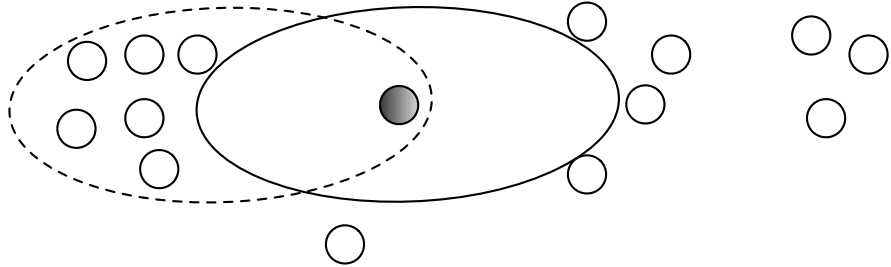}}
  \caption{The behavior of the sum heuristic vs. that of the progress heuristic.  The solid-line
  ellipse indicates the area covered by a search using the sum heuristic.  The dotted-line
  ellipse marks the area searched using the progress heuristic.}
\label{sum-vs-progress}
\end{figure}
Even when there are a few distinct groups of goals in different
directions, the sum heuristic may lead to simultaneous progress
towards all the groups. The two groups of goal states shown in
Figure \ref{sum-vs-progress} illustrate this problem. The sum
heuristic strategy will work only if there are enough enough
resources to reach all the groups. If, however, the resources are
not sufficient, the sum heuristic may waste all the available
resources in trying to progress towards all the groups, but
reaching none.

To avoid such problems, we define the \emph{progress} heuristic,
which takes into account the number of goals towards which
progress is made and the average distance to them. Thus, instead
of trying to pursue multiple groups of goals, this heuristic will
pursue one group at a time, preferring less distant groups.

Let $S_{open}$ be the set of currently opened states.
As before, we
assume
 that we have the explicit goal list,
$S_g=\langle g_1, \ldots , g_k\rangle$, and a distance-estimation
heuristic, $h_{dist}$.
Let $ m_i = \min_{s \in S_{open}}h_{dist}(s,g_{i})$ be
the minimal estimated distance from the search frontier to the
goal $g_{i}$.
For each $s \in S_{open}$ we define
 $G_p(s) = \left \{ g_i \in S_g \: | \: h_{dist}(s, g_i) =
m_i \right \} $ to be the set of goals for which
$s$ is estimated to be the closest among the states in the search frontier.
The average estimated distance between $s$ and the states in $G_{p}$ is
 $D_{p}(s)=\frac{\sum_{g\in
G_p(s)}h_{dist}(s,g)}{\left |G_p(s) \right |}$ be their
average distance from $s$.
We are interested in states that have many members in $G_{p}$
with small average distance.  Hence, we define
 the \emph{progress} heuristic
to be
\begin{equation}
h_{progress}(s)=\frac{D_{p}(s)}{\left |G_p(s) \right |} .
\label{prog-with-sum}
\end{equation}
Minimizing this heuristic will direct the search to larger and
closer groups of goals towards which progress is made. For
example, in the simple space shown in Figure
\ref{sum-vs-progress}, the progress heuristic will advance
correctly, towards the group at the left side as indicated by the
dashed ellipse.  Note that although the right group is larger, the
progress heuristic nonetheless prefers the left one because of its
smaller distance. Since the progress heuristic considers each goal
exactly once, it is not misled by multiple paths to the same goal.

\subsection{Marginal-Utility Heuristics}
\label{musection}
\begin{figure}[!htb]
\centering \scalebox{0.4}{\includegraphics{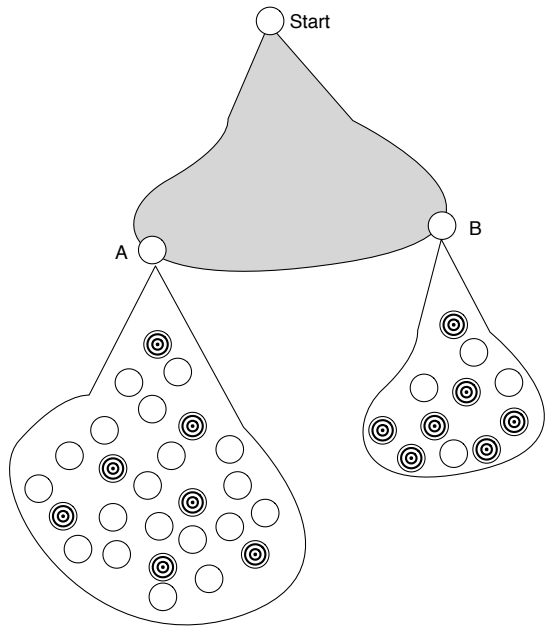}}
\caption{Searching from node $A$ will result in the same number of
goals as searching from node $B$.  It will consume, however, far
greater resources.} \label{marginal-utilityNEEDED}
\end{figure}

 When comparing two search directions, we have so far considered only the concentration of goal
states, preferring directions that lead to larger groups of goals.
Thus, the former heuristics would have considered node $A$ and
node $B$  in Figure \ref{marginal-utilityNEEDED} to be equivalent.
This reasoning, however, does not take into account the resources
invested to reach the set of goals. In the example of Figure
\ref{marginal-utilityNEEDED}, it is clear that visiting the set of
goals under node $B$ requires less search resources than those
under node $A$. Therefore node $B$ is preferable.

We account for these cases by suggesting another approach for
multiple-goal heuristics, one that considers not only the expected
\emph{benefit} of the search but also the expected \emph{cost}.
Obviously, we prefer subgraphs where the cost is low and the
benefit is high. In other words, we would like a high return for
our resource investment.  We call a heuristic that tries to
estimate this return a \emph{marginal-utility} heuristic.

Assume (for now)
that the search space $\langle S, E \rangle$ is a tree.
Let $T(s)$ be the set of all states that are reachable from
$s$.  Let $T_g(s)=T(s) \cap S_g$ be the set of all goal
states reachable from $s$.
Let $S_v \subseteq S$ be the set of all states visited
during a completed search process.
  We define the \emph{marginal utility}
of state $s \in S_v$ with respect to $S_v$ as
\begin{equation}
\mbox{MU}(s)=\frac{\left | T_g(s) \cap S_v \right |}{\left | T(s) \cap
S_v \right |}. \label{original-mu}
\end{equation}
Thus, MU measures the density of goal states in the subtree
expanded by the search process.

One possible good search strategy is to select states that are
should eventually yield high values of $MU$ with respect to $S_v$.
 If the
search process has consumed $R_c \leq R$ resources so far, then
the largest tree that can be visited is of size $r=R-R_c$. In
Section \ref{sec-perfect} we define the perfect heuristic by
considering all the possible ways of distributing $r$ among the
open nodes.  This approach is obviously impractical, and we will
take here a greedy approach instead. We look for a heuristic
function $h_{MU}(s,r)$ that tries to estimate the best marginal
utility of $s$, assuming that all the remaining resources, $r$,
are consumed while exploring $T(s)$. Let $T(s,r)$ be the set of
all trees of size $r$ under root $s$. $h_{MU}(s,r)$ tries to
estimate the \emph{resource-bounded marginal utility}, defined as
\begin{equation}
\label{rb-mu} \mbox{MU}(s,r)=\max_{T \in T(s,r)}\frac{\left | T_g(s) \cap T \right |}{r}.
\end{equation}
$\mbox{MU}(s,r)$ measures the best ratio between the number of
goals achieved and the search resources used for it. Naturally, it
is very difficult to build heuristics that estimate marginal
utility accurately. In the following sections we show how such
heuristics can be learned.

\subsection{Additional Considerations}
\label{secdisab}
 One possible side-effect of not stopping
when discovering goals is the continuous influence of the already
discovered goals on the search process.  The found goals continue
to attract the search front, where it would have been preferable
for the search to progress towards undiscovered goals. If an
explicit set of goal states is given - as it is for the sum and
progress heuristics - we can disable the influence of visited
goals by simply removing them from the set. If a set of features
over states is given instead, we can reduce the effect of visited
goals by preferring nodes that are farther from them in the
feature space. Specifically, let $d(n)$ be the minimal distance of
$n$ from members in the set of visited goals, and let $h(n)$ be
the multiple-goal heuristic value of $n$. The modified heuristic
will be $h'(n)=h(n)(1+c_{1}e^{-c_{2}d(n)})$ where $c_{1}$ and $c_{2}$
are parameters that determine the magnitude of the effect of
$d(n)$. The second term is the penalty we add to the heuristic
value. This penalty decays exponentially with the distance from
the visited goals.

Note that there is a tension between the tendency to search dense
groups of goals and and the tendency to push the search away from
visited goals. When the groups of goals are dense, the above
method can be detrimental because finding some of the goals in a
group reduces the tendency to pursue the other goals of the same
group.  This effect can be controlled by the $c_{i}$ parameters.
For domains with high goal density, $c_{i}$ should be set to lower
values.  Hence, these values can be set dynamically during the
search, according to measurements of goal density in the explored
graph.

 One problem with using the marginal utility
heuristic in non-tree graphs is the possible overlap of marginal
utility.  That means that our search algorithm might pursue the
same set of goals from different directions. One way to overcome
this problem is to try to diversify the search progress by
measuring the feature-based average distance between the ``best''
nodes and the set of recently expanded nodes, and prefer those
with maximal diversity from the nodes explored. This gives maximal
diversity in exploration directions and should minimize the
expected overlap of visited subtrees.

\section{Learning Marginal Utility Heuristics}
\label{sec-learning-mu}
 While it is quite possible that
marginal-utility heuristics will be supplied by the user, in many
domains such heuristics are very difficult to design. We can use a
learning approach to acquire such marginal-utility heuristics
online during the search. We present here two alternative methods
for inferring marginal utility.
One approach estimates the marginal utility of a node based on
the partial marginal utility of its siblings.  The other approach
predicts the marginal utility using
feature-based induction.
\subsection{Inferring Marginal Utility Based on Marginal Utility of Siblings}
\label{SEC:sibling}
The first approach for predicting marginal utility is based
on the assumption that sibling nodes have similar marginal utility.
We define the \emph{partial} marginal utility of a state $s$ at step $t$
of executing a multiple-goal search algorithm as the number of
goals found so far in the subtree below $s$ divided by the number
of states visited so far in this subtree.  Thus, if $S_v(t)$ is
the set of states visited up to step $t$, then the partial
marginal utility is defined as
\begin{equation}
MU(s,t)=\frac{\left | T_g(s) \cap S_v(t) \right |}{\left | T(s) \cap
S_v(t) \right |}. \label{partial-mu}
\end{equation}
The method estimates the marginal utility of the siblings based
on their partial marginal utility, and the marginal utility of the node
based on the average estimated marginal utility of the siblings.

As discussed in the previous subsection, the expected marginal
utility of a node strongly depends on the resources invested in
exploring it.  Thus, to learn the heuristic $h_{MU}(s,r)$, one
would need to estimate partial marginal utility values for
different values of $r$. One way of reducing the complexity of
this two-dimensional estimation is to divide it into two stages:
estimating the depth of the tree under $s$ that is searchable
within $r$ resources, and estimating the marginal utility for the
predicted depth.  Thus, if $\widetilde{MU}_{depth}(s,d)$ is the
estimated marginal utility when searching node $s$ to depth $d$,
we compute the estimated marginal utility of a node $s$ using $r$
resources by
$\widetilde{MU}_{resources}(s,r)=\widetilde{MU}_{depth}(s,\widetilde{d}(s,r))$,
where $\widetilde{d}(s,r)$ is the estimated depth when searching
under node $s$ using $r$ resources. In the following subsections
we show how these values are estimated.

\subsubsection{Updating Partial Marginal-Utility Values}
We maintain for each node two vectors of $D$ counters, where $D$
is a parameter that limits the maximal lookahead of the partial
marginal utility.  One vector, $N(n)$, stores the current number
of visited nodes for node $n$ for depth $1, \ldots, D$, where
$N_{i}(n)$ contains the current number of visited nodes under $n$
with depth of less or equal to $i$.  The other vector, $G(n)$
holds similarly the number of visited goals.

Whenever a new node $n$ is generated, the appropriate entries of
its ancestors' $N$ vectors are incremented. If $n$ is a goal, then
the $G$ vectors are updated as well. If $p$ is an ancestor of $n$
connected to it with a path of length $l \leq D$, then $N_{l}(p),
\ldots N_{D}(p)$ are incremented by one. If more than one path
exists between $n$ and $p$, we consider only the shortest one. The
memory requirements for this procedure are linear in $D$ and in
the number of stored nodes.

The number of operations required for one marginal-utility update
is bounded by $O(B_{degree}^{D})$, where $B_{degree}$ is the upper
bound on the maximum indegree in the graph. Therefore, if the
backward degree is bounded, the number of calculations per node
does not grow as the search progresses. The depth limit $D$
determines the complexity of the update for a given search graph;
hence, it is desirable to reduce its value. A value that is too
low, however, will make it possible to infer only ``local''
marginal-utility values.

\subsubsection{Inferring Marginal Utility}
\label{sec-mu}

The inference algorithm estimates the marginal utility of a node
on the basis of the average partial marginal utility of its
siblings. Only nodes with sufficient statistics about partial
marginal utility are used to predict the marginal utility of new
nodes. We call such nodes \emph{supported} nodes. If the node has
no supported siblings, we base our estimate on the average
estimated marginal utility of its parents (computed recursively
using the same procedure).
 \begin{figure}[!htb]
\centering
\scalebox{0.7}{\includegraphics{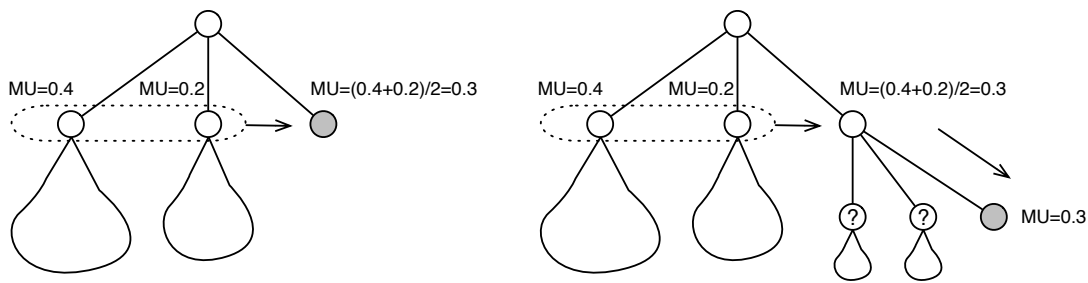}}
\caption{Inferring marginal utility from partial marginal utility
of supported siblings (left) or supported uncles (right). Nodes
with a question mark are unsupported.}
\label{marginal-utilityINFER}
\end{figure}
 Figure
\ref{marginal-utilityINFER} illustrates this method.
In the left tree, the marginal utility of the grey node is computed as
the average marginal utility of its siblings.  In the right tree, the marginal utility of
the grey node is computed as the average marginal utility of its uncles.
The input to the marginal utility estimation heuristic is the
remaining unconsumed resources, $r$. The algorithm first finds the
largest depth, $d$, for which the number of predicted nodes is
smaller than $r$.  This prediction is based on the supported
 siblings
or the parent of the node just as described above.

The found depth, $d$, is then used to determine which counters
will be used to estimate the marginal utilities of the supported
uncles. The complete algorithm is listed in Figure
\ref{m-marginal-utility-calc-val}.
\begin{figure}[!htb]
\begin{center}
\fbox{
\begin{minipage}{\textwidth}
\begin{tabbing}
\TABLINE{0.7cm}
\textbf{procedure} MU($s$,$d$)\\
\> \textbf{if} Supported($s$,$d$) \textbf{then} \textbf{return} $\frac{G_{d}(s)}{N_{d}(s)}$ \\
\> \textbf{else} $P \gets \mbox{Parents}(s)$\\
\> \> \textbf{if} $|P|=0$ \textbf{then} \textbf{return} 0\\
\> \> $\mathit{SupportedSiblings} \leftarrow\{c\in
\mbox{Children}(p)\:|\:p \in P, \mbox{Supported}(c)\}$\\
\> \>\textbf{if} $|\mathit{SupportedSiblings}|>0$ \textbf{then}\\
\>  \>  \>\textbf{return} $\mbox{Avg}\left(\left\{\left.\frac{G_{d}(c)}{N_{d}(c)}\:\right|\:c \in
\mathit{SupportedSiblings}\right\}\right)$\\
\> \>\textbf{else} \textbf{return} $\mbox{Avg}(\{MU(p,Min(d+1,D)) \:|\: p \in P\})$\\
\\
\textbf{procedure} TreeSize($s$,$d$)\\
\> $P \gets \mbox{Parents}(s)$\\
\> \textbf{if} \mbox{Supported}($s$,$d$) \textbf{or} $|P|=0$ \textbf{then} \textbf{return} $N_{d}(s)$ \\
\> \textbf{else}\\
\> \> $\mathit{SupportedSiblings} \leftarrow\{c\in
\mbox{Children}(p)\:|\:p\in P, \mbox{Supported}(c)\}$\\
\> \>\textbf{if} $|\mathit{SupportedSiblings}|>0$ \textbf{then}\\
\>  \>  \>\textbf{return} $\mbox{Avg}(\{\{N_{d}(c)\:|\:c \in
\mathit{SupportedSiblings}\})$\\
\> \>\textbf{else} \textbf{return} $\mbox{Avg}\left(\left\{\left.\frac{\mbox{TreeSize}(p,\mbox{Min}(d+1,D))}{\left|\mbox{Children}(p)\right|} \:\right|\: p \in P\right\}\right)$\\
\\

\TABLINE{0.7cm}\\
\textbf{procedure} Get-marginal-utility($s$,$\mathit{ResourceLimit}$)\\
\>  $\mathit{Depth} = max(d \leq D|\mathit{TreeSize}(s,d)<\mathit{ResourceLimit})$\\
\> \textbf{return} $\mbox{MU}(s,\mathit{Depth})$\\
\end{tabbing}
\end{minipage}}
\end{center}
\caption{\label{m-marginal-utility-calc-val}  An algorithm for
marginal-utility estimation}
\end{figure}
\subsubsection{Sibling Clustering}
\label{sect-sibling}

In the algorithm described in Figure
\ref{m-marginal-utility-calc-val}, the marginal utility of a node
is induced by averaging the partial marginal utility of its
siblings.  If we can define a ``meaningful'' similarity metric
between nodes, we can try making this prediction less noisy by
using only the node's \emph{similar} siblings.
 One way of doing so is to use the similarity metric to cluster
the set of siblings and then generate a virtual node for each
cluster. The virtual node is linked to the cluster members and its
parent is the parent of the original sibling set. This is the only
required change. The existing algorithm described in Figure
\ref{m-marginal-utility-calc-val} will do the rest. When
predicting the marginal utility for a node, the algorithm first
looks for the partial marginal utility of its siblings.  In this
case these are the members of its cluster. Only if these siblings
are unsupported will the algorithm use the information from the
other clusters propagated through the common parent.

\begin{figure}[!htb]
\centering \scalebox{0.5}{\includegraphics{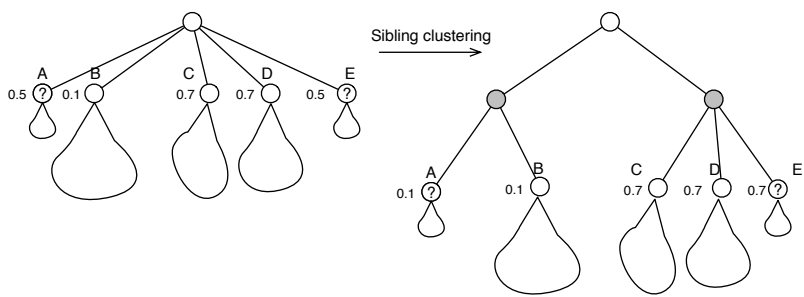}}
\caption{The effect of sibling clustering on marginal utility
estimation} \label{OPSYM}
\end{figure}

This mechanism is illustrated in Figure \ref{OPSYM}. Without
clustering, the predicted marginal utility of the unsupported
nodes would have been the average of the three supported siblings,
which is $0.5$.  Note that this average has a large variance
associated with it. Clustering nodes $A$ and $B$ under one virtual
node, and $C$, $D$, and $E$ under another, yields (we hope) more
accurate prediction, since it is based on more uniform sets.
The similarity metric will usually be  the Euclidean distance
between vectors of features of the states corresponding to the
sibling nodes. In some domains, we try to reduce the number of
generated nodes by deciding at each step which node-operator pair
to proceed with.  In such cases we need a similarity measurement
between operators to implement the above approach.


\subsection{Feature-Based Induction of Marginal Utility}
\label{sec-feat}
 Unfortunately, partial marginal-utility information allows us to predict marginal
utility only for nodes that are in proximity to one another in the
graph structure. In addition, there are domains where the local
uniformity of sibling nodes with respect to marginal utility
cannot be assumed. We can overcome these problems if we view
marginal-utility inference as a problem of function learning and
use common induction algorithms.  For each depth $d$, we induce
the marginal utility function using the set of supported nodes
(with respect to $d$) as examples.  The state features can be
domain independent (such as the in-degree and out-degree of the
node) or domain specific, supplied by the user. As for any
induction problem, the quality of the induced function is highly
dependent on the quality of the supplied features.

Since the above learning scheme is performed on-line, he high cost
of learning, and of using the classifier directly, reduce the
utility of the learning process.  One way to essentially eliminate
the learning costs is to use a \emph{lazy} learner, such as KNN
\cite{Cover:1967:NNP}.  This approach also has the advantage of
being incremental: each new example contributes immediately to the
learned model.  The problem with this approach is the high costs
associated with using the classifier.

An alternative approach would be to learn an efficient classifier
such as a regression tree~\cite{Breiman:1984}.  This can be
learned incrementally using algorithms such as ID5~\cite{Utgoff:1988}.
Batch learning algorithms such as C4.5
usually yield better classifiers than incremental algorithms.
However, due to the higher cost of applying batch learning, one
should decide how often to call it. Applying it after each node
generation would increase the induction cost, but yield better
classifiers earlier - which may improve the performance of the
search process. Applying it at large intervals would reduce the
induction costs but lead to poorer search performance.

 The on-line learning process gives rise to another problem: the initial
 search period where there are not yet sufficient examples to make learning helpful. One way to reduce the effect of
this lack of knowledge is by using classifiers that were induced
beforehand, on-line or off-line, for goals that are similar to the
goals of the current search.

\section{Empirical Evaluation}
\label{sec-exps}
 To test the effectiveness of the methods described in the previous
sections and to show their versatility, we experimented intensively
on several domains.
The most challenging domain, however, is \emph{focused crawling} where
we apply our algorithms to the task of collecting target web pages
from a sub-web of millions of pages.
We first compare the anytime behavior of our distance-based
methods with that of uninformed search and best first search.
Then we test the performance of our marginal utility methods.
We also test the effect of the various suggested enhancements
on the algorithms' performance.   We also show
realtime performance of our algorithm by allowing it
to search the real web.
\subsection{Experimental Methodology}
We compare the performance of our algorithms to two
competitors: breadth-first search and best-first search using
distance estimation heuristics. Both algorithms were adopted to
the multiple-goal framework by allowing them to continue the
search after finding the first goal.

A basic experiment that compares two multiple-goal search
algorithms is conducted in the following way:
\begin{enumerate}
  \item A set of initial states and a goal predicate are given.
  \item The algorithms perform multiple-goal search.
  \item The resources consumed and the number of goals found during the execution are monitored.
  \item The last two steps are repeated several times
  to accumulate sufficient statistics (all the algorithms
  contain at least one random component).
  \item The performance of the two algorithms, measured by the number of goals found for the allocated resources, is compared.
 \end{enumerate}

The problem of estimating the performance of an algorithm when
a resource
allocation is given is most extensively discussed in the context of
anytime algorithms.
 Measuring the performance of anytime algorithms
is problematic~\cite{hansen96monitoring}. If a probability
distribution over the resource allocation is given, then we can
compute the expected performance of an anytime algorithm on the
basis of its performance profile. In many cases, however, such a
probability distribution is not available. We therefore measure
the performance of the tested algorithm by means of the obtained
quality for different resource allocation values.
 For multiple-goal search, the quality is measured by the
number of goals found for the allocated resources. When we know
the total number of goals, we report instead the percentage of
goals found.

Alternatively, anytime algorithms can be evaluated by measuring
the amount of resources required to achieve a given quality.  For
multiple-goal search, the most obvious measurement is time. Time,
however, is overly affected by
    irrelevant factors such as hardware, software, and
    programming quality.  Moreover, in the Web domain,
    most of it is spent accessing Web pages. How long this takes depends on many factors, such as network and server loads,
    which are irrelevant to our research topic.

We thus decided to measure resource consumption by the number of
generated nodes. Nevertheless, we cannot ignore time completely:
we must make sure that the overhead of the methods described in
this paper does not outweigh their benefits. We therefore report
time results for an experiment that uses the real Web.

Many parameters affect the performance of the algorithms described
in this paper.  Ideally, we would like to perform \emph{factorial
analysis}~\cite{Montgomery:2001} so that each combination of
values is tested.  Such experimentation, however, is infeasible
for the large number of variables involved. We therefore take the
one-factor-at-a-time approach, where we use a default value for
all parameters except the one being tested. In addition, wherever
appropriate, we perform several experiments testing two factors
together.


\subsection{Tasks And Domains}
\label{sec_tasks}
  Most of our experiments are conducted in the context
of several Web domains. To show the generality of our approach, we
  applied our methods to several additional domains, including
 n-queens, open knight tours, multiple robot path planning and
 multiple sequence alignment.

 Our algorithms were applied to the following tasks:
\begin{enumerate}
  \item{\em Focused crawling:}
    One of the main motivations for this research is the problem of
    \emph{focused crawling} in the Web
    \cite{chakrabarti99focused,cho00evolution,Kleinberg:1997,menczer01evaluating}.
    The task is to find as many goal pages as possible using limited
    resources, where the basic resource unit is usually  the actual retrieval of a page from a link.
While it looks as if the task of retrieval information from
internet could have been achieved using general-purpose search
engines, there are several circumstances where focused crawling is
still needed:
\begin{enumerate}
\item When the search criterion is complicated and is not
expressible in the query language of search engines. \item When
one needs an updated set of goals -- search engines are updated
every few weeks due to the huge space the brute-force crawlers
have to cover. \item When the coverage of the general engines is
not sufficient.
\end{enumerate}

    Previous work on focused crawling concentrated
    on Web-specific techniques for directing the search.  Our experiments
    will test whether our generalization of single-goal heuristic search to
    multiple-goal search can contribute to the task of focused
    crawling.

    Performing rigorous empirical research on the Web is problematic.
    First, the Web is dynamic and
    therefore is likely to be modified between different
     runs of the algorithms~\cite{douglis97rate}.
    Second, the enormous time required for crawling in the Web disallows parametric experimentation. To solve the above problems we
    downloaded a significant section of the Web to local storage and
    performed the experiments using the local copy~\cite{cho98efficient,hirai99webbase}. Specifically, we downloaded a large
     part of the {\tt
    .edu} domain, containing, after some cleanup, approximately
    8,000,000 valid and accessible HTML pages.  The resulting graph
    has an average branching factor of 10.6 (hyperlinks).

     We tested the performance of the algorithms on the entire downloaded domain.
     Some of the parametric experiments, however, were too time-consuming even when the local copy was used.
     Therefore, we used small sub-domains for these experiments.
     A sub-domain is generated by randomly selecting a root page out of a predesignated
     set of roots and extracting a sub-graph of size 35,000 under it.

     To ensure that the overhead of our algorithms does not
     significantly affect their performance, we also conducted several
     online experiments on the real Web.

We use three types of goal predicates:
\begin{enumerate}
\item Predicates that test for pages about specific
    topics: robotics, mathematics, football, food, and sport.
    The predicates were automatically induced by applying decision tree
    learning to a set of manually supplied examples.
\item Predicates that test for certain \emph{types} of pages:
    pages containing publication lists, laboratory pages, student home pages,
    project pages and news pages.  These predicates were also learned from examples.
\item Predicates that test for home pages of people who are
members of a specific list. Each list of people was generated from
a Web page listed the names of people together some personal
information (such as name, affiliation and area of interest). The
predicates employ commonly used heuristics for determining whether
an HTML document is a home page of a specific person.  We use
three such predicates corresponding to three different lists we
found on the Web.  We limit each list to contain 100 names.
\end{enumerate}
    \item{\em Finding paths to multiple goals:}
    Path-finding algorithms usually search for a single path
    to one of the goals.  In some applications, however,
    we get a set of initial states and a set of goal states and our
    task is to find a \emph{set} of paths from an initial state to a goal state.
    These paths may then be used by another algorithm that
    evaluates them and selects one to execute according to
    various criteria.  We simulated a physical environment
    by a
    $500 \times 500$ grid with random ``walls'' inserted as obstacles (an average of 210000 nodes with an average 3.9 branching factor).
     There are parameters controlling the maximal length of
     walls and the desired density of the grid.
    Walls are inserted randomly while making certain that the resulting graph remains connected.
    The set of goal states is randomly generated using one of the
    following two methods:
    \begin{enumerate}
    \item A set of states is independently and uniformly drawn from the
    set of all states.
    \item 10\% of the goal states are generated as above.  We then randomly and
    uniformly select $K$ states that are used as cluster centers.
    The rest of the goal states are then randomly generated with distances
    from the centers that are normally distributed.
    \end{enumerate}
    Figure \ref{GRID} shows an example
    of a multiple path-finding problem and a solution that
    includes 9 paths.
   \begin{figure}[!htb] \centering
\scalebox{0.5}{\includegraphics{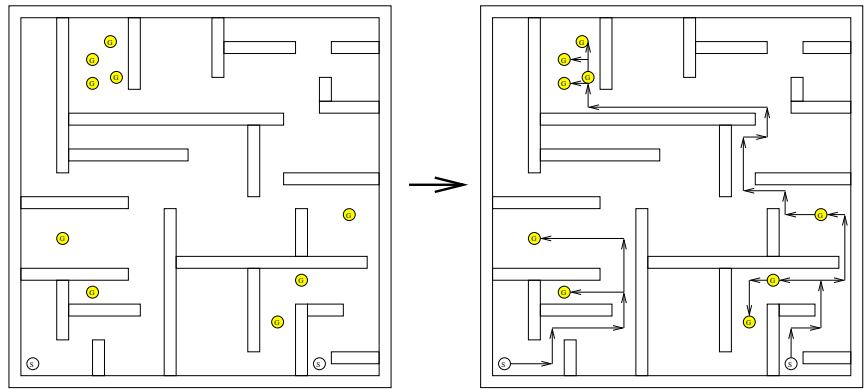}}
\caption{Searching for a set of paths to multiple goals in a
grid}\label{GRID}
\end{figure}
    \item{\em Planning movement of multiple robots:}
   Assume that we are given a set of $N$ robots located in various
   states, while their task is to collect a set of $K>N$ objects scattered
   around.  In this case we need to plan $N$ paths that will pass
   through as many objects as possible.  This situation is illustrated
   in Figure \ref{GRIDR}.  Although the problem appears to resemble the one in Figure \ref{GRID},
   the solution here has two paths (while the solution to the previous problem has 9).
      \begin{figure}[!htb] \centering
    \scalebox{0.5}{\includegraphics{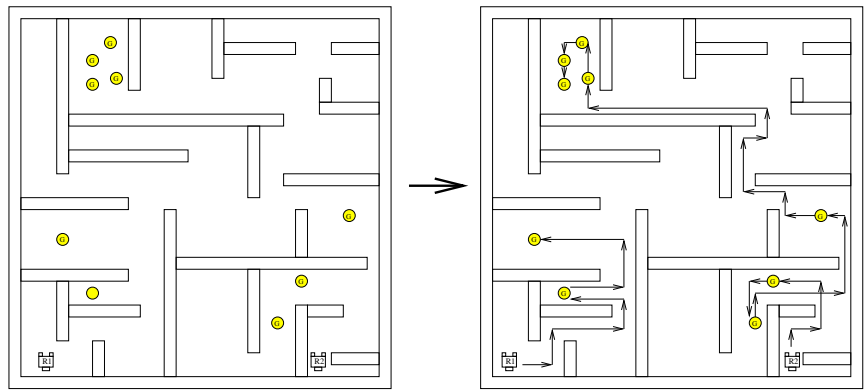}}
    \caption{Multiple-robot path
    planning in a grid}\label{GRIDR}
    \end{figure}
    There are many similar planning problems:  for example,
      that of planning product delivery
    from several starting points to multiple customers using a fixed
    number of delivery trucks.
    For the experiments here we used the same type of grids as for the previous problem.
    The robots were placed randomly.  We assumed that collisions are not harmful.
    \item{\em Open knight tours:}
      This is a famous problem where the task is to find a path
      for a knight on a chess board such that each of the
      squares is visited and none is visited twice.  We tried
      a multiple-goal version of it where the task is to find
      as many such paths as possible within the allocated
      resources.
   For our experiments, we
    used boards of $6 \times 6$, $7 \times 7$ and $8 \times 8$ squares.
\item{\em N-Queens:} A constraint satisfaction problem where
 the goal is to place N
queens on a chessboard such that no two queens are in the
 same row, column or diagonal. In the multiple goal version of this
problem, we want to find as many satisfying configurations as
possible within the allocated resources.
 \item{\em Multiple sequence alignment:}
A known bioinformatics problem where the goal is to align several
biological sequences optimally with respect to a given cost
function. In the multiple-goal version we are interested in
obtaining as many almost optimal solutions as possible within the
allocated resources.
\end{enumerate}


\subsection{Performance with Distance-Based Heuristics}
We experimented first on the two multiple-goal heuristic functions
that are based on graph-distance estimation: the sum heuristic and
the progress heuristic. We compare the performance of
multiple-goal best-first search that uses these heuristics with:
\begin{enumerate}
\item Multiple-goal best-first search that uses the distance
estimation as its heuristic function. \item Breadth-first search
(BFS) (shown by~\citeR{Najork:2001}, to perform
  well for Web crawling).
\end{enumerate}

The sum heuristic requires distance estimates to individual goals.
We define such distances for three of our domains. For the
multiple path finding and multiple robot planning we use the
Manhattan distance.  For focused crawling task we experiment with
the personal home page domain. We estimate the distance between a
given page and the home page of a list member by computing the
cosine vector distance between the bag of words of the given page
and the bag of words of the description text for the person.

The distance heuristic and the sum heuristic were tested with and
without disabling the influence of visited goals.  We did not use
disabling with the progress heuristic since it subsumes this
behavior.  For the two grid-based tasks, where all the goals are
given, we used complete disabling by removing the visited goals
from the goal list as described in Section \ref{secdisab}.

For the personal home page search task, where the set of goals is
not explicitly given, we used the feature-based method described
in Section \ref{secdisab}. The features used for determining the
distance between candidate pages and visited goals are the words
with the highest \emph{TFIDF} value~\cite{joachims97probabilistic,Salton:1988:TWA}.

\begin{figure}[htb]
  \twofig{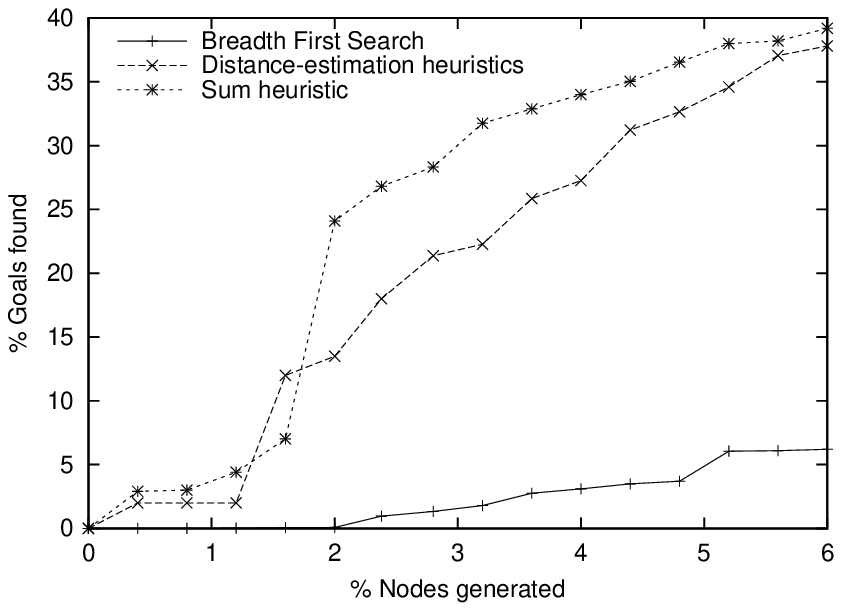}{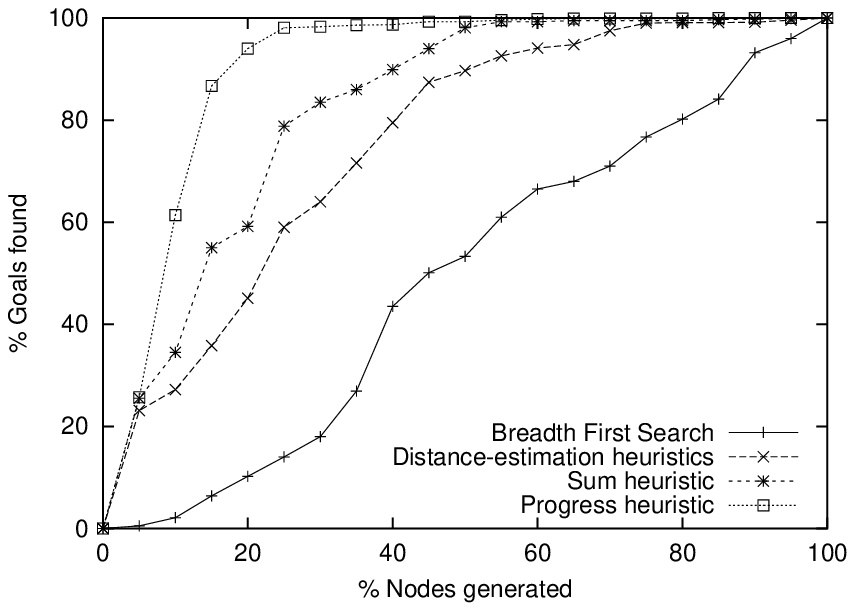}
  \caption{ Anytime performance of the various heuristics:
            (a)Focused crawling  (b)Multiple path finding}\label{heuristic-anytime}
\end{figure}

\begin{table}[t]
\small
\begin{center}
  \begin{tabular}{||c|c||c||c|c||c|c||c||}
 \hline

    Domain& Task & BFS & \multicolumn{2}{c||}{Min. dist.} & \multicolumn{2}{c||}{Sum} & Progress  \\
    \cline{4-7}
    &&& Without & With  & Without & With &\\
    &&& disab. & disab. & disab &  disab &\\\hline\hline
         \multicolumn{2}{||c||}{} & \multicolumn{6}{c||}{\% of goals found at 20\% resources}  \\
\hline

    Multiple path & scattered    & $8.5\PM0.1$  & $21.3\PM1.2$ & $29.0\PM0.5$ & $21.5\PM0.9$ & $28.9\PM0.3$& $76.8\PM2.1$\\
    \cline{2-8}
    finding & clustered          & $10.2\PM0.1$ & $34.0\PM0.5$ & $45.1\PM0.4$ & $32.6\PM1.1$ & $59.2\PM0.3$ & $94\PM1.2$  \\
    \hline\hline
    Multiple robot   & scattered & $7.1\PM0.8$  & $20.3\PM0.8$ & $26.5\PM0.4$& $22.3\PM0.6$ & $25.8\PM0.2$& $89.9\PM1.8$\\
    \cline{2-8}
    movement         & clustered & $10.1\PM0.9$ & $31.2\PM1.4$ & $47.4\PM1.2$& $42.0\PM0.6$ & $64.1\PM0.7$ & $98.6\PM0.9$\\
    \hline
    \hline
    \multicolumn{2}{||c||}{}& \multicolumn{6}{c||}{\% of goals found at 2\% resources} \\
    \hline
    Focused & Group 1        & $0.1\PM0.0$ & $13.5\PM0.5$ & $17.3\PM1.3$ & $24.1\PM0.7$& $28.0\PM1.1$& $51.3\PM0.8$\\
    \cline{2-8}
     crawling   & Group 2        & $3.2\PM1.4$ & $18.4\PM2.1$ & $26.2\PM1.9$ & $19.7\PM1.0$& $23.5\PM1.1$& $78.1\PM3.1$\\
    \cline{2-8}
    100-person search                  &Group 3        & $0.3\PM0.1$ & $5.8\PM0.9$    & $10.7\PM1.4$ & $6.4\PM0.9$ & $11.7\PM0.9$& $60.9\PM0.8$\\
    \hline
  \end{tabular}
  \caption{\small The performance of multiple-goal search with various heuristics.  The numbers in parentheses are standard deviations.}\label{table-heuristics}
\end{center}
\end{table}
Table \ref{table-heuristics} summarizes the results of this
experiment.  Each number represents the average of 50 experiments.
We measure the performance after consuming 20\% of the maximal
amount of resources, i.e., expanding 20\% of the total number of
nodes in the search graph. The 100-person home page search proved
to be a relatively easy domain where our methods were able to find
most of the goals after consuming very little resources.
Therefore, for this domain, we measure the performance at 2\% of
the nodes. Figure \ref{heuristic-anytime} shows the anytime
behavior of the various methods for two of the domains.  The
graphs for the other domains show similar patterns.

On the basis of this table and corresponding graphs, we make the
following observations:
\begin{enumerate}
\item The progress heuristic is superior to all other methods
tested so far.  That it is advantageous for the case of clustered
goals comes as no surprise because it leads to one cluster being
pursued at a time. That it is superior to distance estimation for
the case of scattered goals is far less obvious: we would expect
both heuristics to pursue one goal after another and therefore
yield similar results. The weighted progress heuristic, however,
prefers pursuing goals that are closer to other goals, thus
yielding better results. \item   The results clearly demonstrate
that the goal influence phenomenon is indeed significant, and our
method is efficient in reducing this effect. \item In almost every
case, heuristic methods are significantly better
 than blind search.  The only exception is when using the sum heuristic
 without influence disabling in graphs with scattered goals.  In such
 cases, its behavior is only marginally better than blind search.

\end{enumerate}

\subsection{Performance with Marginal-Utility Heuristics}
\label{secinf}
 None of the methods in the previous subsection take into account the
expected search resources involved in pursuing the alternative
directions. In addition, they all assume either knowledge of the
specific set of goals or of the heuristic distances to each.
 In this subsection we test the performance of the
two methods that are based on marginal utility as described in Section
\ref{sec-mu}. The experiments described in this subsection are
performed for the focused crawling task with the 10 goal
predicates described in Section \ref{sec_tasks}.

For the topic-based goals, we cannot compare the performance of
the marginal-utility algorithms to that of the sum heuristic
because we do not have access to the list of goals or to a list of
heuristic values for specific goals as in the previously tested
domains. Therefore we use for comparison blind BFS and the common
best-first search using the distance-estimation heuristic. The
heuristic is based on the list of words selected by the induction
algorithm when generating the goal predicates.
\subsubsection{Inferring Marginal Utility from Partial Marginal Utility}
\label{sec-muexp}
 We implemented and tested the marginal-utility
estimation algorithm described in Section \ref{SEC:sibling}.
\begin{figure} [!ht]
  \centering
  \twofig{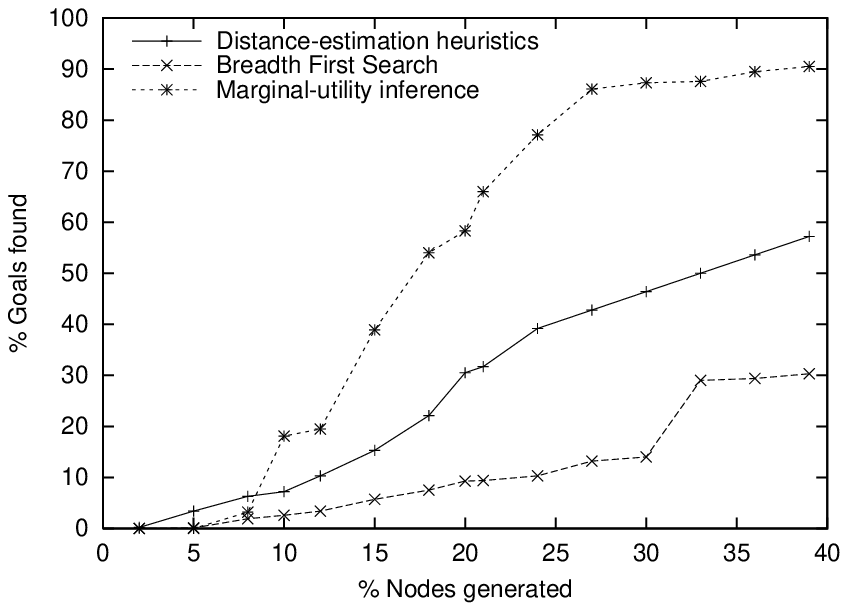}{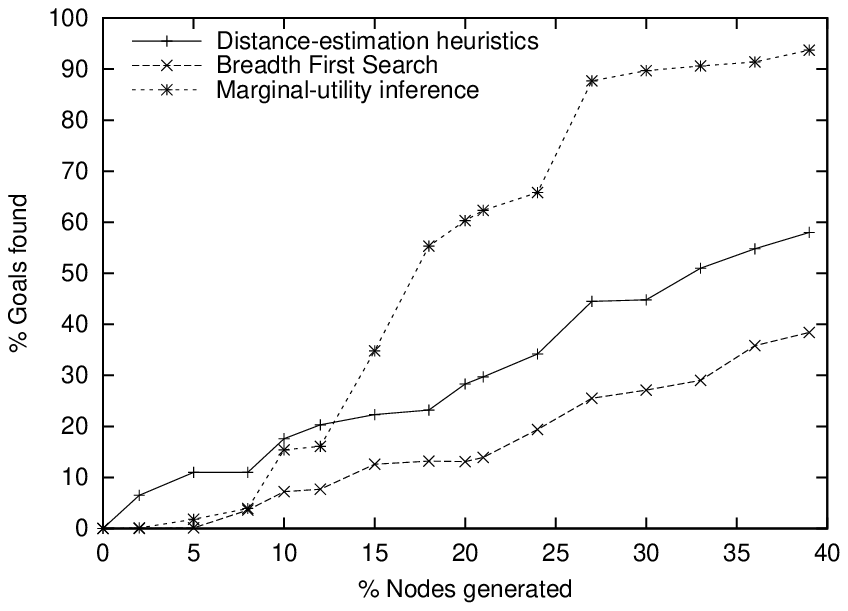}
  \caption{ The performance of multiple-goal best-first search
using the marginal-utility inference method applied to focused
crawling (with D=4).  The results are shown for (a)Robotics pages
(b) Mathematics pages}\label{web-heuristic}
\end{figure}
Figure \ref{web-heuristic} shows the performance profiles of the
tested algorithms for the \emph{robotics} and \emph{mathematics}
goal predicates. Each graph represents the average of 5 runs of
the tested algorithm using 5 starting pages randomly selected from
the fixed 200 root pages. In both cases we see a significant
advantage of the marginal-utility method over the other two
methods.  This advantage becomes evident only after an initial
``training'' period, in which sufficient statistics are
accumulated.  The full data for the 10 domains with resource
allocation of 10\% and 20\% is available in the Appendix.
\begin{figure} [ht]
  \begin{center}
  \scalebox{0.9}{\includegraphics{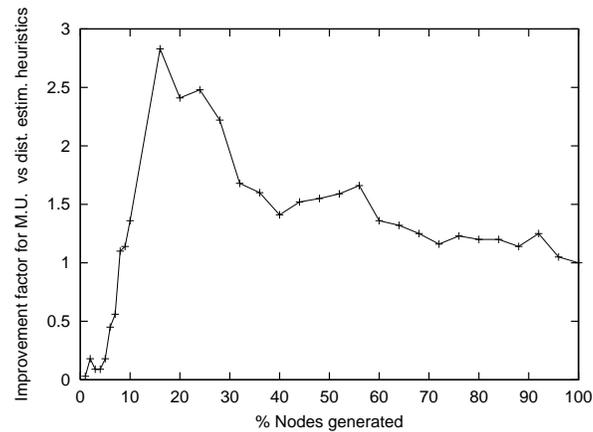}}
  \end{center}
  \caption{ The improvement factor for best-first using marginal-utility
inference compared to best-first using the distance-estimation
heuristic}\label{imp-factor}
\end{figure}
 Figure
\ref{imp-factor} shows the average improvement factor (compared to
distance estimation)
for the 10 domains as a
function of the search resources. The graph shows very nicely the
initial ``exploratory'' stage where the statistics are accumulated
 until about $8\%$ of the resources have been consumed, after which
 the improvement factor becomes larger than $1$. The improvement
 factor reaches a peak of about $2.8$
at $17\%$, and then starts to decline towards a value of $1$ at
$100\%$ search resources,
 where any algorithm
necessarily finds all the goals.

 Performance at the exploratory stage can be improved by combining the two methods.
 We tested a hybrid method, which
uses a linear combination of the marginal-utility prediction and
the heuristic estimation. To determine the linear coefficients of
each part, we conducted 100 experiments on the small Web subgraph
(below 35,000 pages) using different goal predicates (unrelated to
those tested in our main experiments).
\begin{figure} [!ht]
  \centering
  \twofig{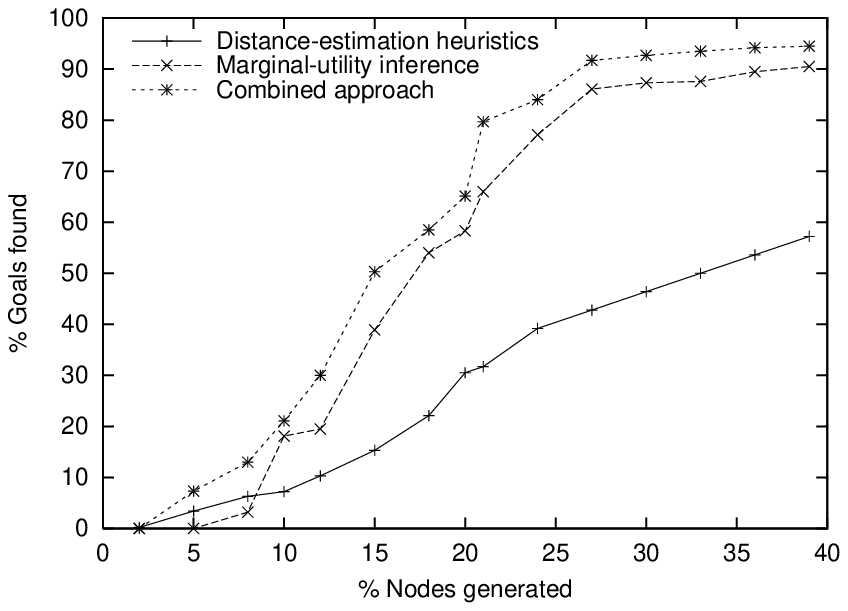}{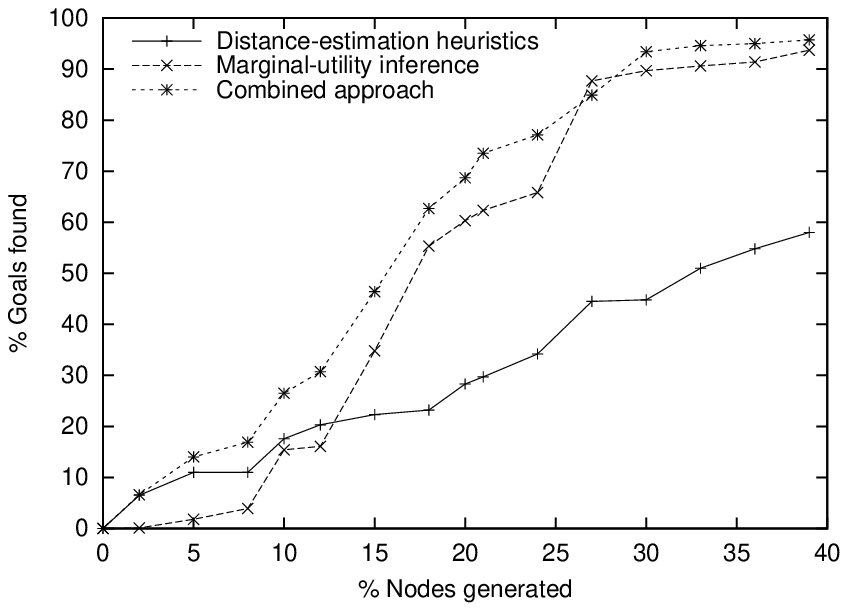}
  \caption{ Combining marginal-utility inference with heuristic search for (a)Robotics pages (b)Mathematics pages.}\label{combined-approach}
\end{figure}
Figure \ref{combined-approach} shows the results obtained for the
combined method compared to each of the individual methods.
 We can see that indeed the combined method is better
than each of the algorithms alone. An interesting phenomenon is
that, at all points, the result for the combined method is better
than the maximal results
 for the other two.  One possible explanation
  is that very early in
the search process the distance-estimation heuristic
  leads to a sufficient number of goals to jump-start the learning process much earlier.
  The results for the 10 domains are available in the Appendix.
\subsubsection{Learning Marginal-Utility from Features}
\label{sec-learn} In Section \ref{sec-learning-mu} we describe a
method for feature-based generalization of visited search nodes in
order to induce marginal-utility values.  We conducted a set of
experiments to test the efficiency of the learning mechanism for
the problem of focused crawling with the same 10 goal types as in
previous experiments.

During crawling, we accumulate the \emph{supported} visited pages
and tag them with their marginal utility as measured at the time
learning took place (see Section \ref{sec-learning-mu}). We then
convert the tagged pages to feature vectors and hand them to the
CART algorithm~\cite{Breiman:1984} for regression-tree
induction\footnote{Other induction algorithms, such as SVM or
Naive Bayes, could have been used as well.}. We then used the
induced tree to estimate the marginal utility of newly generated
 nodes.

For features, we use the bag-of-words approach, which is dominant
in the field of text categorization and classification. The value
of each word-feature is its appearance frequency.  Words appearing
in HTML title tags are given more weight.  We apply feature
selection to choose words with the highest \emph{TFIDF}~\cite{joachims97probabilistic,Salton:1988:TWA}.
\begin{figure} [!ht]
  \twofig{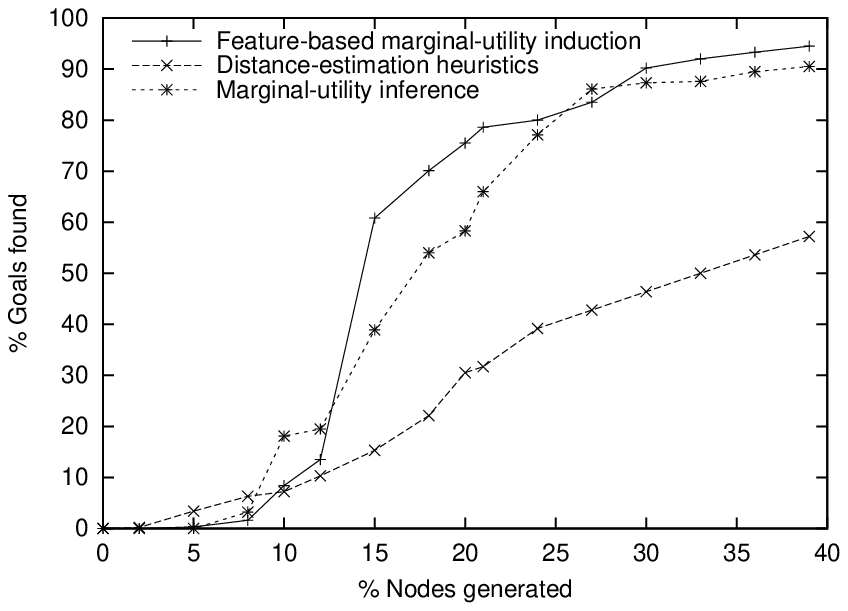}{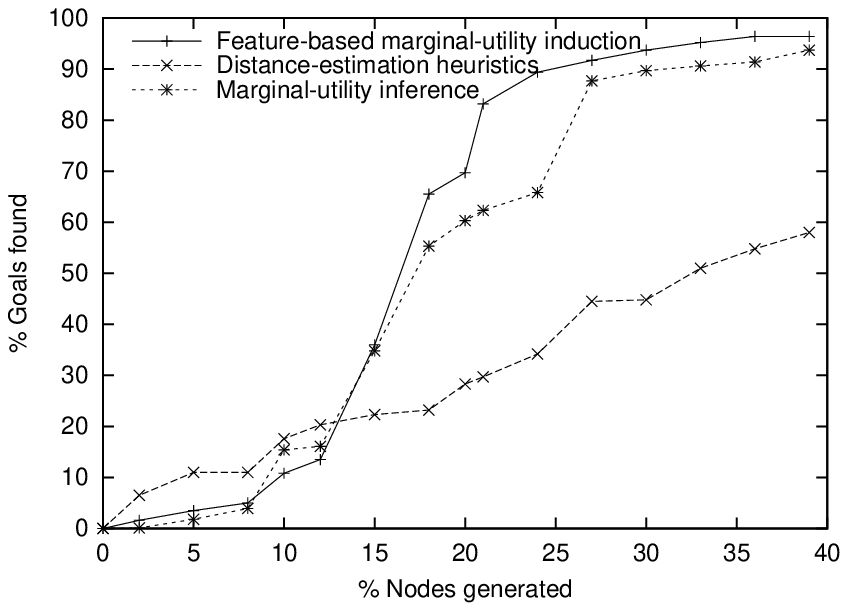}
  \centering
  \caption{The performance of best-first search with marginal utility
  induced using the regression trees classifier.
  The experiments were performed on the problem of focused crawling with:
  (a)Robotics pages (b) Mathematics pages}\label{learnvsinfer}
\end{figure}

Figure \ref{learnvsinfer} shows the results obtained for the
\emph{robotics} and \emph{mathematics} goal predicates. The full
report for the 10 domains is available in the Appendix. In
both cases, there is an initial period where the sibling-based
inference method outperforms the more sophisticated feature-based
induction. After this period, however, the induction-based method
significantly outperforms the sibling-based method. One possible
reason for the initial inferior performance is that feature-based
induction requires more examples than the simplistic sibling-based
method that only computes averages and therefore needs fewer
examples.

We inspected the produced trees and found out that
they reflect reasonable concepts.  They have several
dozens of nodes and contain both features that are
related to the searched goal and features that correspond
to hubs (such as \emph{repository} and \emph{collection}).

We have tested whether our choice of classifier affects the
performance of the induction-based method by performing the same
set of experiments using the KNN classifier. The results obtained
were essentially identical.

\subsection{Testing the Full System Architecture}
\label{SEC:full-system} 
\begin{figure} [!ht]
  \twofig{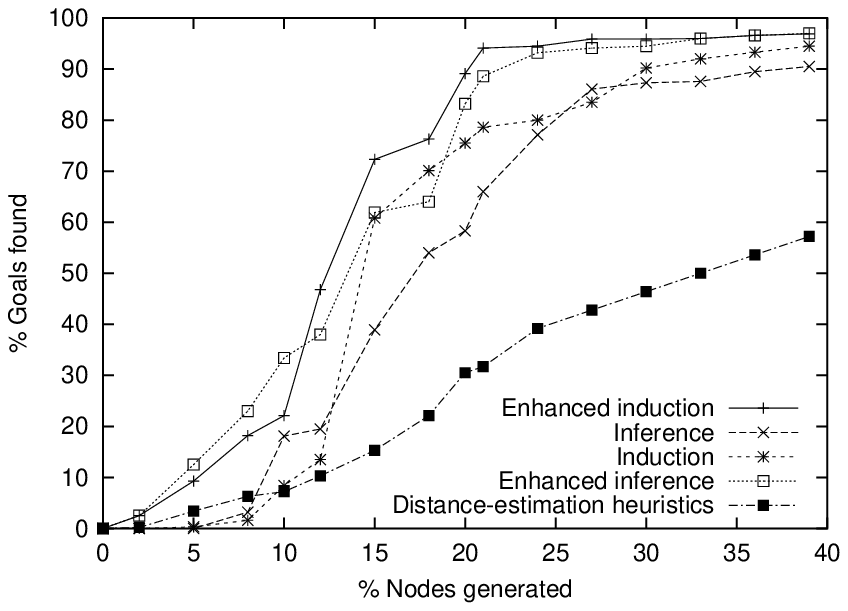}{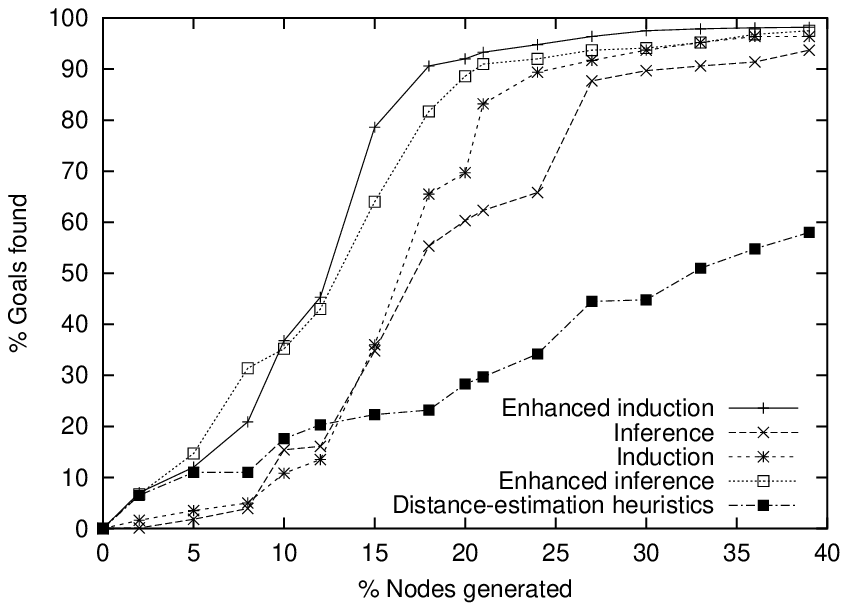}
  \centering
  \caption{The performance of best-first search (using different marginal-utility
  heuristics) with all the enhancements enabled:
  (a)Robotics pages (b) Mathematics pages.
  Each figure contains 5 plots: one for the baseline performance
(\emph{distance estimation}),  two for the unenhanced methods (\emph{inference}
and \emph{induction}) and two for the enhanced methods (\emph{enhanced inference}
and \emph{enhanced induction}).
\label{finallearn}}
\end{figure}

We have described various enhancements of
our marginal-utility methods, including sibling clustering,
overlap minimization, combining the marginal-utility heuristic
with the distance-estimation heuristic, and disabling of visited
goals. Figure \ref{finallearn} shows the performance of
 the two marginal-utility methods with the
full set of enhancements.
The full results for the 10 domains are
available in the Appendix.

 Both the sibling- and feature-based methods indeed improve when all the enhancements are enabled.
Furthermore, the feature-based method maintains its advantage,
albeit with a slightly decreased magnitude.
\begin{figure} [!ht]
  \twofig{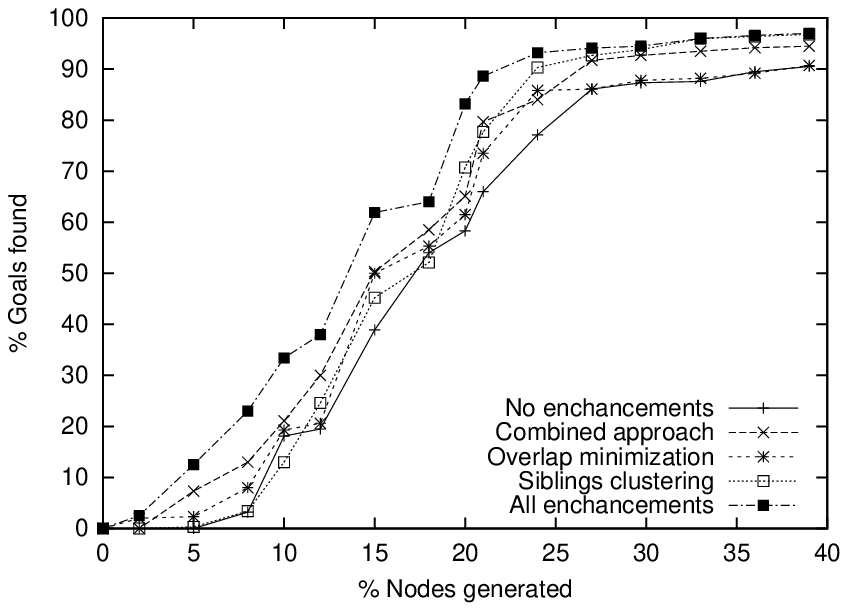}{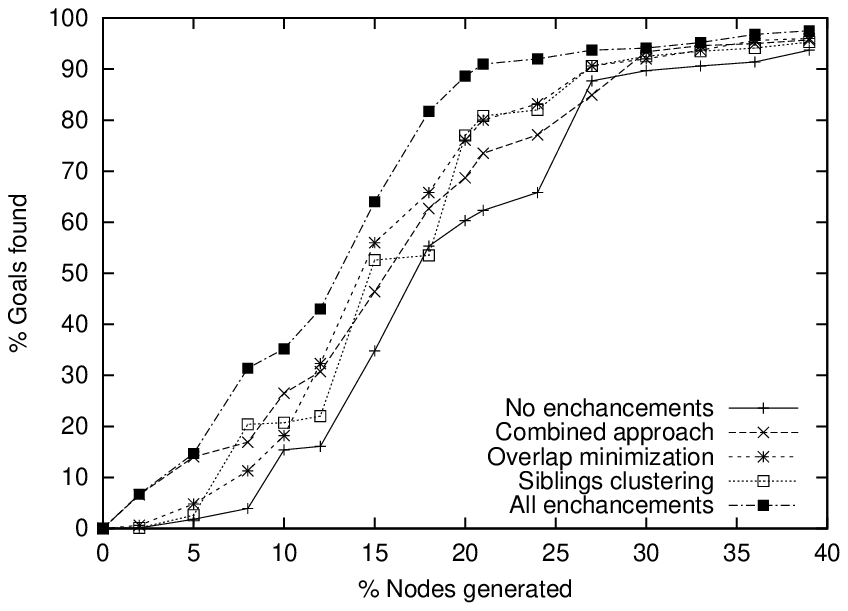}
  \centering
  \caption{The performance of best-first search (using marginal-utility inference) with all the enhancements
   enabled
  compared to the performance of the algorithm with a single option enabled:
  (a)Robotics pages (b) Mathematics pages.
  Each figure contains 5 plots: One for the marginal utility inference method with no enhancements,
  one for the same method enhanced by the combined approach, one for enhancement by overlap estimation, one for enhancement by sibling clustering and, finally, one for all the enhancements together.
  \label{finallearn-infer}}
\end{figure}
Although enabling all the enhancements does improve system
performance, it should be recalled that the same is true for
enabling each enhancement separately. A question thus arises
whether the improvements are at least partially cumulative. In
other words, would performance using \emph{all} the enhancements
be better than performance using each of the enhancements
separately? Figure \ref{finallearn-infer} compares all the graphs
for the sibling-based method.  We can see that the fully enhanced
method is indeed superior to all the rest.
\subsection{Realtime Performance}

In the previous experiments we took the number of generated nodes
as a basic resource unit.  We must be careful, however, since this
measurement does not take into account the overhead in our method.
To ensure that the overhead does not outweigh the benefits, we
conducted a realtime evaluation of the system architecture
performing focused crawling on the online Web and measured the
resources consumed in the time that elapsed.\footnote{ The
experiment was performed using a Pentium-4 2.53 GHz computer
 with 1GB of main memory and a cable modem.
}.
\begin{figure} [!ht]
  \centering
  \twofig{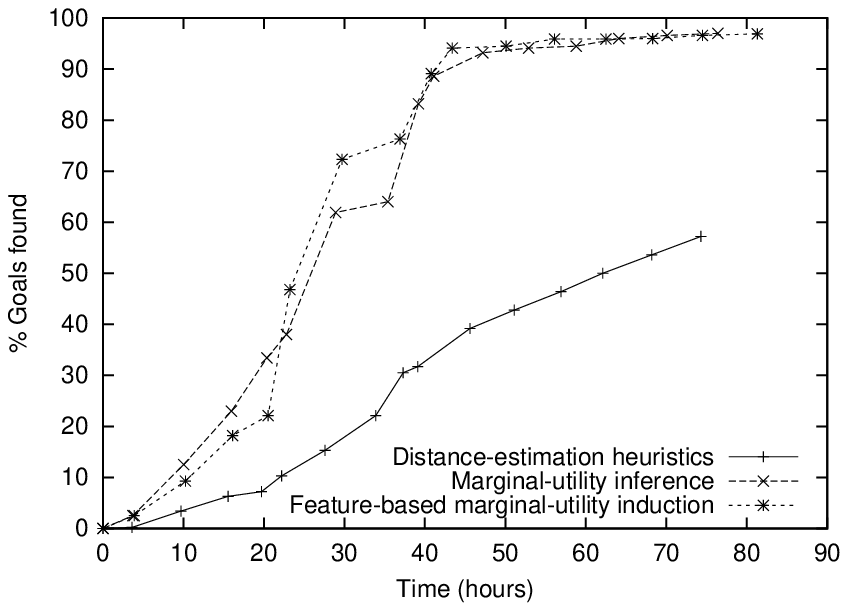}{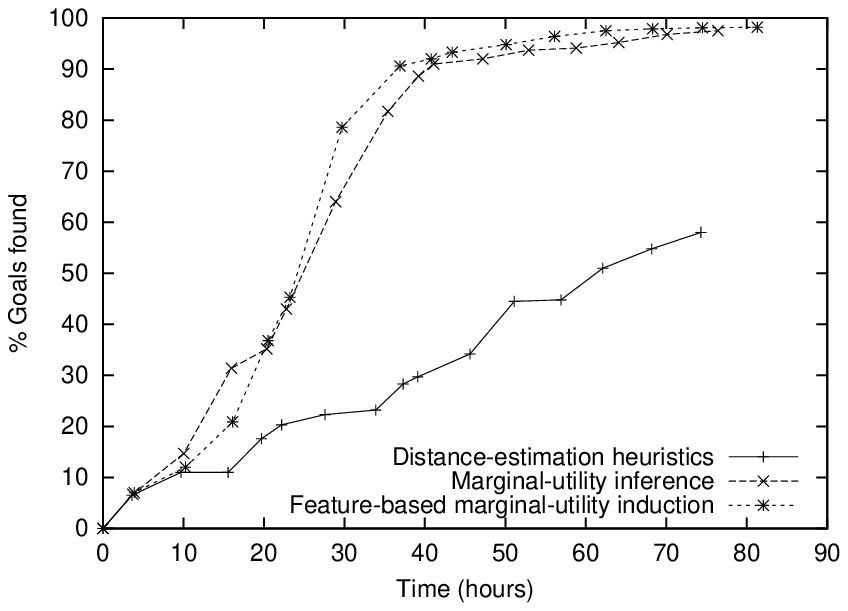}
  \caption{The performance of the marginal-utility based methods
   as a function of real time: (a)Robotics pages (b)Mathematics
  pages}
  \label{realtimeinf}
\end{figure}
Figures \ref{realtimeinf}(a),(b) show the performance of our
methods
 with all the enhancements enabled as a function of real time.
 A comparison of these graphs to the graphs shown in Figure
 \ref{finallearn} reveals that the overhead of our methods does not noticeably affect their performance.
 \begin{figure} [!ht]
  \centering
  \scalebox{0.9}{\includegraphics{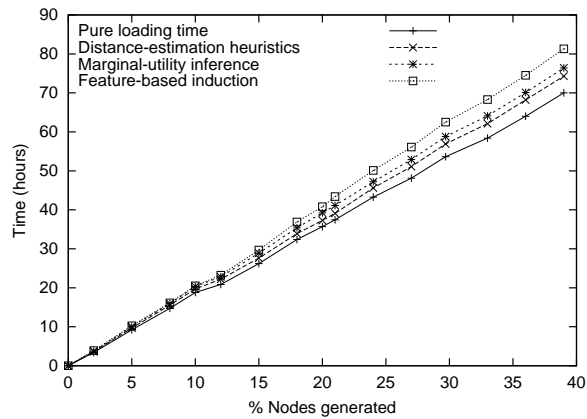}}
  \caption{The average real time used by the marginal-utility methods
   as a function of the number of generated nodes in the focused crawling domain}
  \label{realtime}
\end{figure}
To see if the overhead increases with the size of the visited
graph, we plotted in Figure \ref{realtime} the real time as a
function of the number of generated nodes. The graphs show
 that the majority of the time required for focused crawling tasks is indeed the loading time
itself, even when all calculations related to the discussed
options are enabled. In fact, using distance-estimation increases
the computation time by a factor of no more than 1.07; using
similarity-based inference of marginal utility, by no more than
1.1; and using feature-based induction, by no more than 1.17.
Thus, if our improvement is greater than these factors, our
algorithms will benefit focused crawling systems.

\subsection{Contract Algorithms Versus Interruptible Anytime Algorithms}
 \label{sec-contract}
 The experiments described so far test the performance of
 our methods as interruptible anytime algorithms.
 Each point of the graph can also be considered as
 a test result for a contract algorithm using the
 specific resource allocation.  However,
 if a multiple-goal search algorithm is called in
 contract mode, we can utilize the additional
 input (of resource allocation) to improve the
 performance of our algorithm, as described
 in Section \ref{sec-mu}.

 To test the effect of exploiting the resource allocation,
 we repeated the experiment described in Section \ref{sec-muexp}
 using the algorithm of Section \ref{sec-mu}.  We found out that
 when using the algorithm in contract mode, we obtained
  an average improvement factor of $7.6$ (factor of $2.8$ if dropping two extremes) for 5\%
  resource allocation and $1.4$ for 10\% resource allocation.  The full results
  are available in the Appendix.

 Our algorithms also allow a ``contract by quality'' mode where the
input is the required quality instead of the allocated resources.
In our case the quality is specified by the percentage of goals
found.  The contract algorithm achieved an average improvement
factor of $1.9$ for 5\% of the goals and $1.4$ for 20\% of the
goals. The full results are available in the Appendix.

\subsection{The Performance of Other Multiple-Goal Search Algorithms}
In previous subsections we test our heuristic methods on
best-first search. To show the generality of our approach, we test
whether marginal-utility heuristics can be used efficiently for
dynamic ordering of variables in backtracking, and for choosing a
node from the focal group in the multiple-goal $A^*_\epsilon$
algorithm. In both cases we used the sibling clustering method.

For backtracking we used the multiple-goal version of two known
problems: open knight tour and n-queens. We applied our
marginal-utility inference algorithm which based on similarity of
siblings to sort variable values in the multiple-goal version of
backtracking search. For the open knight tour problem we used a
$6\times6$ board (this board contains $524,486$ goal
configurations, and this is the maximal size where we can collect
all the goals efficiently). For the n-queens problem we used a
$16\times16$ board containing $14,772,512$ goals. In neither case
did we use a domain-specific technique to increase search
efficiency.
\begin{figure} [!ht]
  \twofig{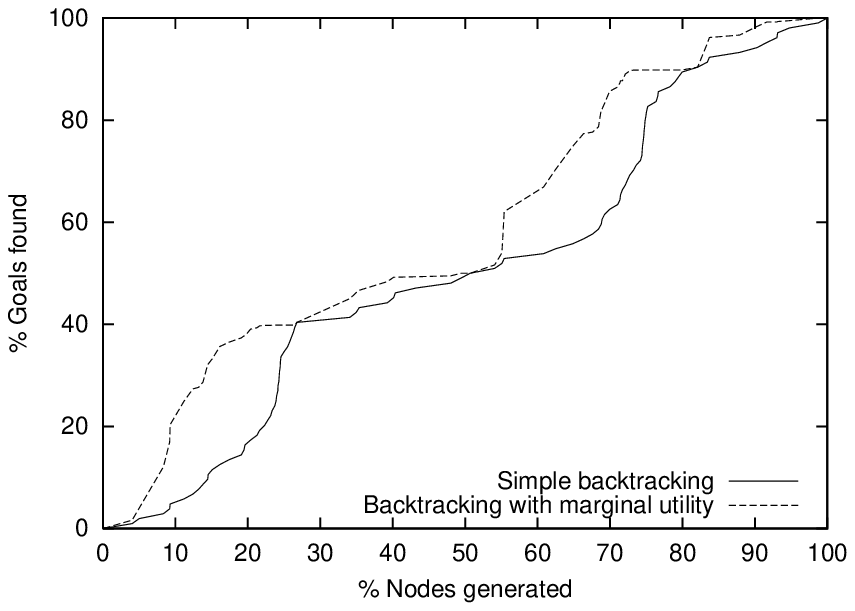}{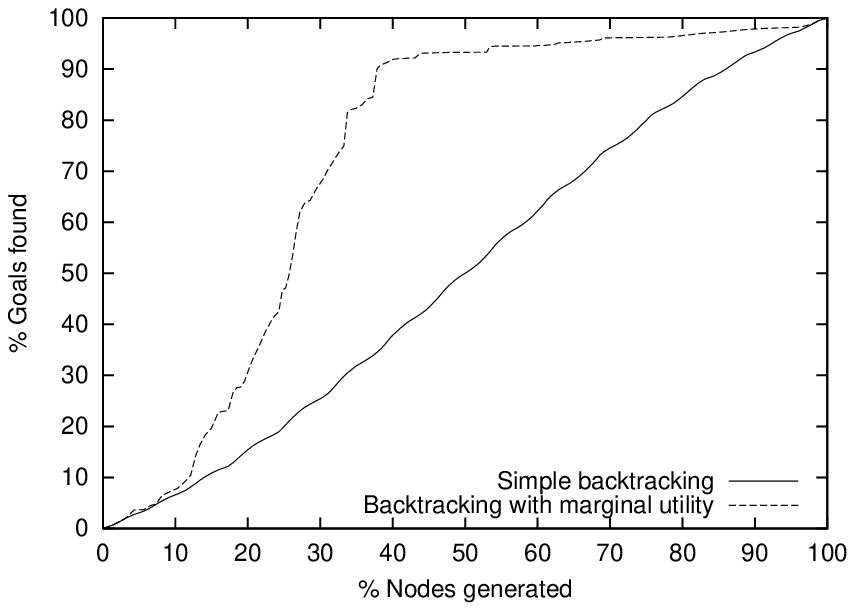}
  \centering
  \caption{Applying marginal-utility inference to backtracking search
  (a)Open knight tour (b) N-queens task}\label{backtrack}
\end{figure}
Figure \ref{backtrack}(a),(b) compares the performance of
multiple-goal backtracking search with and without marginal
utility.  We can see that applying marginal-utility inference
significantly increases the performance for both problems.

We also tested the multiple-goal $A^*_\epsilon$ algorithm,
described in Section \ref{searchalgs}, where the sibling-based
marginal-utility heuristic selects a node from the focal group.
The experiment was performed on the known multiple-sequence
alignment problem. Since the graph degree is very large, we
applied a modified version of $A^*$ described by Yoshizumi, Miura
and Ishida~\citeyear{yoshizumi00with}. We used the same heuristic,
methodology and data set described in this paper, requiring the
algorithm to collect all optimal and 1.1-suboptimal solutions.

\begin{figure} [!ht]
  \twofig{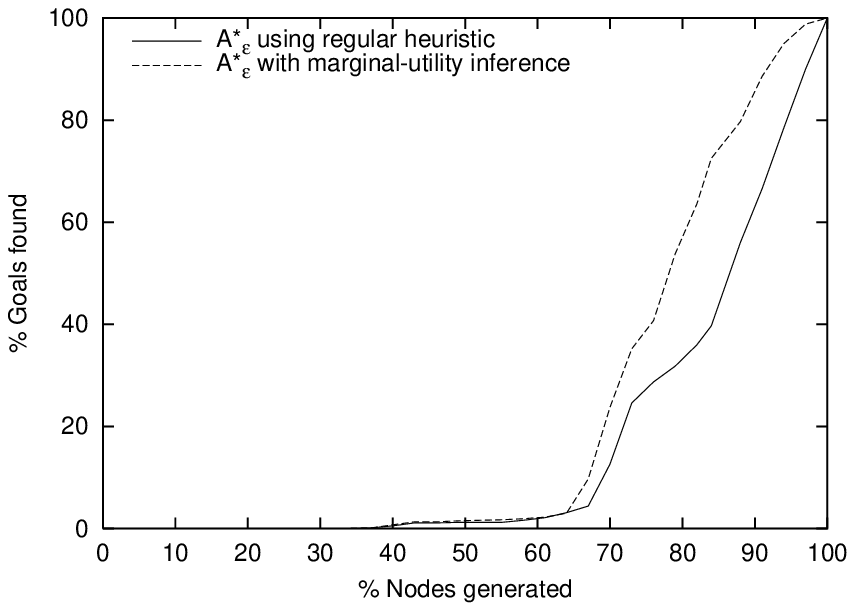}{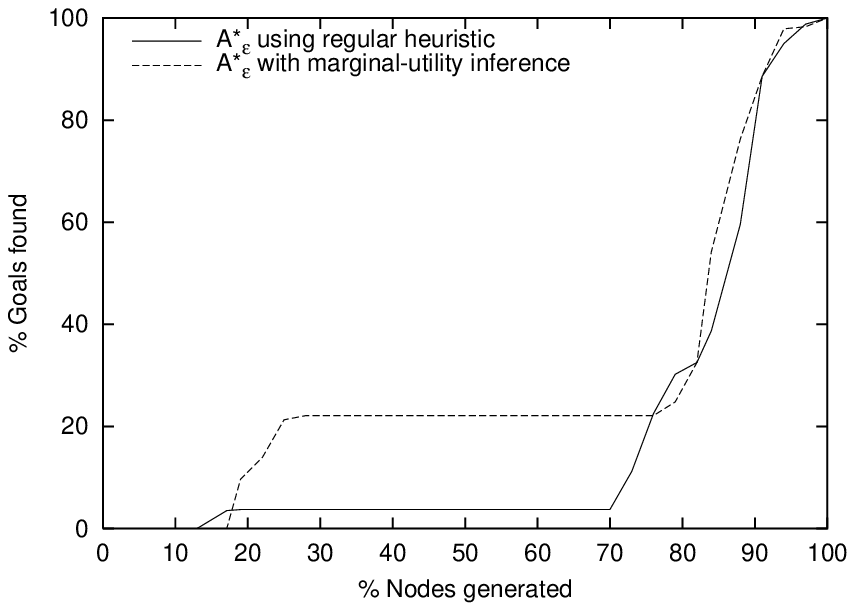}
  \centering
  \caption{Applying marginal-utility inference to $A^*_\epsilon$ search
  (a)Data set 1 (b) Data set 2}\label{astar}
\end{figure}

Figure \ref{astar} shows the results obtained for two data sets.
We can see that marginal utility improves the performance of the
search algorithm.

\section{Related Work}

The multiple-goal search framework defined in this paper is novel.
No previous work has treated heuristic search for multiple goals
as a general search framework and no previous work provided
\emph{general} algorithms for multiple-goal search. The planning
community has dealt with multiple goals only in an entirely
different setup, where the goals are conjunctive and possibly
conflicting. In particular, that setup is not easily applicable to
generic graph search.

Even domain-specific algorithms for multiple-goal heuristic search
are not very common.  Of the domains mentioned in Section
\ref{sec_tasks}, the only one that given any attention as a
multiple-goal problem domain was Web crawling. The sequence
alignment domain was used in several works on heuristic search
\cite<for example,>{korf:2002,Zhou:2002:MSA,Zhou:2003:SAS,Schroedl:2005:ISA}, but
only as a single-goal search problem.

The popularity of the Web has led many researchers to explore the
problem of multiple-goal search in the Web graph. This problem is
better known as focused crawling. \citeA{chakrabarti99focused}
defined a focused crawler as ``a Web agent which selectively seeks
out pages that are relevant to a pre-defined set of topics by
retrieving links from the live Web.'' Such agents can be used for
building domain-specific Web indices \cite<see for
example>{mccallum99building}. Focused Web-crawling algorithms use
various methods such as breadth-first search~\cite{Najork:2001},
best-first search~\cite{cho00evolution,cho98efficient}, and
reinforcement learning~\cite{boyan96machine,rennie99using}. Most
of the heuristic methods for focused crawling are based on
Web-specific features.  For example, the page-rank model
\cite{page98pagerank} was used by \citeA{brin98anatomy}, and by
\citeA{haveliwala99efficient}.  The hubs-and-authorities model
\cite{Kleinberg:1997} was used by \citeA{borodin01finding}. In
addition, several theoretical works provide analysis and bounds
for the problem of Web crawling \cite<for
example>{cooper02crawling,kumar00web}.

The approach most similar to ours is that taken by by Rennie and
McCallum~\cite{rennie99using}, who apply reinforcement learning
to Web crawling.  Like our marginal-utility induction method,
their method also estimates a reward value and generalizes it over
unvisited nodes. There are, however, several important differences
between the two methods, particulary in the definition of reward.
While our approach is based on the maximal number of goals
achieved for the given resources, their method is focused on
immediacy of goal achievement. The sooner a goal can be achieved
by an optimal algorithm starting from a graph node, the more it
contributes to the reward value of this node regardless of whether
the algorithm has enough resources to collect this goal. Thus,
their setup does not allow a direct incorporation of supplied
resource limit input.
 Furthermore, their
approach relies on relatively heavy off-line processing on a
training set. We propose an online update method to estimate and
update the marginal-utility based system. Our approach not only
eliminates the need for fetching the fixed training set, but also
gives more flexibility to the algorithm.

\section{Discussion}

    The work described in this paper presents a new framework for
heuristic search. In this framework the task is to collect as many
goals as possible within the allocated resources.  We show that
the traditional distance-estimation heuristic is not sufficiently
effective for multiple-goal search. We then introduce the
\emph{sum} and \emph{progress} heuristics,
 which take advantage of an explicitly given goal set to estimate
 the direction to larger and closer
groups of goals.

One problem with the above heuristics is that they ignore the
expected resources required to collect goals in the alternative
directions.  We introduce the marginal-utility heuristic, which
attempts to estimate the cost per goal of each search direction.
Thus, using it should lead to a more productive search.

    Designing an effective marginal-utility heuristic is a rather
    difficult task.  We therefore developed two methods for online
learning of marginal-utility heuristics. One is based on local
similarity of the partial marginal-utility of sibling nodes, and
the other generalizes marginal-utility over the state feature
space. Both methods infer marginal utility from the partial
marginal-utility values which are based on the number of visited
goal and non-goal nodes in partially explored subgraphs.

The sibling-based inference method requires only the basic input
of a search problem: a set of starting nodes, a successor
function, and a goal predicate.  The method can also take
advantage of an input resource allocation, as was demonstrated in
Section \ref{sec-contract}. If a distance-estimation heuristic is
given, the sibling-based method can utilize it for the initial
stages of the search where the data on which to base the inference
is not sufficient. The marginal-utility generalization method
requires a set of meaningful features over the set of states. This
is a common requirement for learning systems.

    We applied our methodology to several tasks, including focused
Web crawling, and showed its merit under various conditions.  We
also applied it to the tasks of finding paths to multiple goals,
planning movement of multiple robots, knight-tour, n-queens, and
finding a set of multiple sequence alignments. The experiments
show  that even without any prior knowledge about the goal type,
and given only the goal predicate, our algorithm, after an
initiation period, significantly outperforms both blind and
best-first search using a distance-estimation heuristic. When we
enhance our method with a regular distance-estimation
 heuristic, our method shows more than threefold
improvement over the same distance-estimation heuristic alone.

Our framework and proposed algorithms are applicable
to a wide variety of problems where we
are interested in finding many goal states rather than only one.
We show, for example, for the multiple sequence alignment problem,
that our methods allow  $A^*_\epsilon$ to reach many more
nearly-optimal configurations. The same methods can be also
applied to constraint satisfaction problems where it may be useful
to find many solutions and apply another algorithm for selecting a
solution from the found set.

To apply our framework to new problems, the following requirements
must be fulfilled:
\begin{enumerate}
\item The problem domain should be formulated as a state space.
\item There exists predicate that identifies goal states. \item
For using the sum and progress heuristics:
\begin{enumerate}
\item A traditional function that estimates the distance between two states should be given.
\item The set of goals states should be given explicitly.
\end{enumerate}
\item For the sibling-based method of marginal utility inference
we assume that the marginal utility values of sibling nodes are
relatively similar. \item For the induction-based marginal utility
inference we assume the availability of a set of state features
that are informative with respect to marginal utility.
\end{enumerate}

The main message of this research is that the induction-based
method for marginal utility inference should be used when
possible.  Unlike the sum and the progress heuristics, it takes
into account the resources needed for collecting the goals. Unlike
the sibling-based inference method, it makes no assumptions about
similarity of marginal utility values between siblings.  It does
require informative state features; however, for many domains, a
reach set of state features -- sufficient for inducing the marginal
utility estimation function -- is available.

Our marginal-utility based heuristic techniques are greedy in the
sense that they always choose the node leading to a subgraph where
we would expect to find the maximal number of goals, were we to
use all the remaining resources for that subgraph. A more
sophisticated approach would try to wisely distribute the
remaining resources between promising directions, leading, we
hope, to better performance than the greedy approach.

Although the marginal utility approach does not consider possible
overlap between subgraphs, we propose, in Section \ref{secdisab},
a technique to reduce it. A more interesting direction could be to
predict the actual overlap as the search progresses, using the
same methods as in the partial marginal-utility calculation.

The framework described in this paper opens a new research
direction.  The field is wide open for the development of new
algorithms and for the application of multiple-goal search
algorithms to other tasks.


\section*{Acknowledgements}

 We would like to thank Adam Darlo,
Irena Koifman and Yaron Goren, who helped us in programming. This
research was supported by the fund for the promotion of research
at the Technion and by the Israeli Ministry of Science.



\appendix


\section{Detailed Results}
In this appendix we provide a breakdown of the results for the 10
Web topics.  Tables \ref{marginal-utilitygeneral-table} and
\ref{marginal-utility-combining-table} refer to the results
described in Section 6.4.1. Table \ref{marginal-utility-induction-table}
refers to the results described in Section 6.4.2. 
Table \ref{marginal-utility-final-table}
refers to the results described in Section 6.5.
Tables \ref{limiting-resources} and \ref{limiting-goals}
refers to the results described in Section 6.7.

\begin{table}[ht]
\small
\begin{center}
  \begin{tabular}{|c||c|c|c||c|c|c||}
 \hline
    Goal type  & \multicolumn{3}{c||}{\% of goals found at 10\% resources}& \multicolumn{3}{c||}{\% of goals found at 20\% resources} \\
    \cline{2-7}
    &  BFS & Distance & Marginal & BFS & Distance & Marginal\\
    &      & estimation & utility &     & estimation & utility\\
    \hline
    Robotics                    & $2.6\PM0.9$  & $7.2\PM0.2$   & $18.1\PM0.6$ & $9.3\PM1.6$  & $30.5\PM1.1$& $58.3\PM2.0$\\
    Students                    & $10.5\PM1.0$ & $11.8\PM0.5$  & $17.0\PM1.3$ & $12.5\PM1.5$ & $23.7\PM0.6$& $54.6\PM1.3$\\
    Mathematics                 & $7.2\PM0.1$  & $17.6\PM1.6$  & $15.4\PM0.8$ & $13.1\PM1.2$ & $28.3\PM0.8$& $60.3\PM2.9$\\
    Football                    & $0.0\PM0.0$  & $31.4\PM0.5$  & $25.3\PM0.9$ & $0.8\PM0.0$  & $42.1\PM1.7$& $71.5\PM0.7$\\
    Sports                      & $3.6\PM0.4$  & $37.0\PM1.1$  & $45.1\PM2.7$ & $16.2\PM0.7$ & $41.6\PM0.7$& $68.5\PM1.4$\\
    Laboratories                & $0.3\PM0.1$  & $12.5\PM0.7$  & $33.4\PM0.8$ & $5.3\PM0.5$  & $18.1\PM0.8$& $80.4\PM3.1$\\
    Food                        & $7.2\PM1.2$  & $25.1\PM0.9$  & $30.3\PM1.7$ & $26.5\PM3.9$ & $48.5\PM1.7$& $92.7\PM2.2$\\
    Publications                & $0.3\PM0.1$  & $12.6\PM1.0$  & $35.2\PM2.6$ & $3.5\PM0.1$  & $18.5\PM1.1$& $64.2\PM1.9$\\
    Projects                    & $0.1\PM0.0$  & $30.1\PM1.4$  & $41.2\PM1.6$ & $0.8\PM0.2$  & $32.0\PM1.4$& $80.5\PM2.0$\\
    News                        & $8.5\PM0.9$  & $23.1\PM0.8$  & $22.6\PM0.8$ & $15.3\PM1.0$ & $40.4\PM1.2$& $88.9\PM0.7$\\
    \hline
 \hline
  \end{tabular}
  \caption{\small Marginal-utility and distance-estimation heuristics in focused crawling (with D=4). The numbers in parentheses are standard deviations.}\label{marginal-utilitygeneral-table}
\end{center}
\end{table}

  \begin{table}[!ht]
  \small
\begin{center}
  \begin{tabular}{|c||c|c|c||c|c|c||}
 \hline
    Goal type  & \multicolumn{3}{c||}{\% of goals found at 10\% resources}& \multicolumn{3}{c||}{\% of goals found at 20\% resources}  \\
    \cline{2-7}
    & Distance & Marginal & Combined & Distance & Marginal & Combined \\
    & estimation & utility & method& estimation & utility & method\\
    \hline
    Robotics                    & $7.2\PM0.2$   & $18.1\PM0.6$ & $21.1\PM0.5$& $30.5\PM1.1$& $58.3\PM2.0$ & $65.1\PM2.1$\\
    Students                    & $11.8\PM0.5$  & $17.0\PM1.8$ & $23.3\PM0.6$& $23.7\PM0.6$& $54.6\PM1.3$ & $59.6\PM1.2$\\
    Mathematics                 & $17.6\PM1.6$  & $15.4\PM0.8$ & $26.5\PM1.9$& $28.3\PM0.8$& $60.3\PM2.9$ & $68.7\PM2.5$\\
    Football                    & $31.4\PM0.5$  & $25.3\PM0.9$ & $33.9\PM1.0$& $42.1\PM1.7$& $71.5\PM0.7$ & $70.9\PM1.1$\\
    Sports                      & $37.0\PM1.1$  & $45.1\PM2.7$ & $48.3\PM2.2$& $41.6\PM0.7$& $68.5\PM1.4$ & $75.0\PM1.3$\\
    Laboratories                & $12.5\PM0.7$  & $33.4\PM0.7$ & $40.2\PM0.9$& $18.1\PM0.8$& $80.4\PM3.1$ & $79.8\PM1.0$\\
    Food                        & $25.1\PM0.9$  & $30.3\PM1.7$ & $30.1\PM1.2$& $48.5\PM1.7$& $92.7\PM2.2$ & $93.3\PM1.6$\\
    Publications                & $12.6\PM1.0$  & $35.2\PM2.7$ & $42.0\PM2.4$& $18.5\PM1.1$& $64.2\PM1.9$ & $77.8\PM1.7$\\
    Projects                    & $30.1\PM1.4$  & $41.2\PM1.6$ & $41.3\PM1.0$& $32.0\PM1.4$& $80.5\PM2.0$ & $84.7\PM1.8$\\
    News                        & $23.1\PM0.8$  & $22.6\PM0.8$ & $30.5\PM1.9$& $40.4\PM1.2$& $88.9\PM0.7$ & $90.5\PM0.9$\\
    \hline
 \hline
  \end{tabular}
  \caption{\small Combining marginal-utility with distance-estimation heuristics for focused crawling. The numbers in parentheses are standard deviations.}\label{marginal-utility-combining-table}
\end{center}
\end{table}

  \begin{table}[!ht]
  \small
\begin{center}
  \begin{tabular}{|c||c|c|c||c|c|c||}
 \hline
    Goal type  & \multicolumn{3}{c||}{\% of goals found at 10\% resources}& \multicolumn{3}{c||}{\% of goals found at 20\% resources} \\
    \cline{2-7}
    & Distance & Inference  & Inference    & Distance & Inference  & Inference \\
    & estim. & by siblings & by induction    & estim. & by siblings & by induction\\
    \hline
    Robotics                    & $7.2\PM0.2$   & $18.1\PM0.7$ & $8.4\PM1.2$& $30.5\PM1.1$& $58.3\PM2.0$ & $75.5\PM2.5$\\
    Students                    & $11.8\PM0.6$  & $17.0\PM1.9$ & $12.4\PM1.1$& $23.7\PM0.6$& $54.6\PM1.3$ & $68.2\PM1.6$\\
    Mathematics                 & $17.6\PM1.6$  & $15.4\PM0.9$ & $10.8\PM1.5$& $28.3\PM0.8$& $60.3\PM2.9$ & $69.7\PM2.5$\\
    Football                    & $31.4\PM0.5$  & $25.3\PM0.4$ & $21.1\PM0.7$& $42.1\PM1.7$& $71.5\PM0.7$ & $71.6\PM1.1$\\
    Sports                      & $37.0\PM1.2$  & $45.1\PM2.7$ & $36.2\PM1.6$& $41.6\PM0.7$& $68.5\PM1.4$ & $79.4\PM1.5$\\
    Laboratories                & $12.5\PM0.7$  & $33.4\PM0.8$ & $19.4\PM1.3$& $18.1\PM0.8$& $80.4\PM3.1$ & $86.0\PM1.3$\\
    Food                        & $25.1\PM0.9$  & $30.3\PM1.7$ & $27.1\PM0.6$& $48.5\PM1.7$& $92.7\PM2.2$ & $89.1\PM1.8$\\
    Publications                & $12.6\PM1.0$  & $35.2\PM2.7$ & $21.8\PM2.4$& $18.5\PM1.1$& $64.2\PM1.9$ & $78.9\PM1.2$\\
    Projects                    & $30.1\PM1.4$  & $41.2\PM1.8$ & $35.5\PM1.2$& $32.0\PM1.4$& $80.5\PM2.0$ & $89.8\PM2.5$\\
    News                        & $23.1\PM0.8$  & $22.6\PM0.8$ & $23.2\PM0.9$& $40.4\PM1.2$& $88.9\PM0.7$ & $93.9\PM1.1$\\
    \hline
    \hline
  \end{tabular}
  \caption{\small The performance of the two methods of
   marginal-utility estimation for the focused crawling task.
  The numbers in parentheses are standard deviations.}\label{marginal-utility-induction-table}
\end{center}
\end{table}

  \begin{table}[!ht]
  \small
\begin{center}
  \begin{tabular}{|c||c||c|c||c|c||}
 \hline
    Goal type  & \multicolumn{5}{c||}{\% of goals found at 10\% resources} \\
    \cline{2-6}
    & Distance-estimation & \multicolumn{2}{c||}{Inference by siblings} & \multicolumn{2}{c||}{Inference by induction}\\
    \cline{2-6}
    &       & \small{Without} & \small{With}  & \small{Without}     & \small{With}  \\
    &  &  \small{enhanc.}&\small{enhanc.}&\small{enhanc.}&\small{enhanc.} \\
    \hline
    Robotics                    & $7.2\PM0.2$   & $18.1\PM0.7$ & $33.4\PM2.1$ & $8.4\PM1.2$& $22.1\PM2.2$\\
    Students                    & $11.8\PM0.5$  & $17.0\PM1.8$& $26.5\PM1.0$ & $12.4\PM1.1$& $25.2\PM1.6$\\
    Mathematics                 & $17.6\PM1.6$  & $15.4\PM0.8$& $35.2\PM1.4$ & $10.8\PM1.4$& $36.8\PM1.5$\\
    Football                    & $31.4\PM0.6$  & $25.3\PM0.9$ & $46.9\PM1.4$ & $21.1\PM0.7$& $39.5\PM1.4$\\
    Sports                      & $37.0\PM1.2$  & $45.1\PM2.7$& $54.9\PM1.6$ & $36.2\PM1.6$& $46.8\PM1.5$\\
    Laboratories                & $12.5\PM0.7$  & $33.4\PM0.8$ & $46.4\PM1.6$  & $19.4\PM1.3$& $39.5\PM1.4$\\
    Food                        & $25.1\PM0.9$  & $30.3\PM1.7$ & $35.1\PM1.4$  & $27.1\PM0.6$& $30.4\PM1.4$\\
    Publications                & $12.6\PM1.1$  & $35.2\PM2.6$ & $46.8\PM1.5$ & $21.8\PM2.4$& $30.0\PM1.4$\\
    Projects                    & $30.1\PM1.4$  & $41.2\PM1.6$ & $49.2\PM1.2$ & $35.5\PM1.2$& $44.6\PM1.3$\\
    News                        & $23.1\PM0.8$  & $22.6\PM0.8$ & $41.1\PM1.6$ & $23.2\PM0.9$& $42.3\PM1.1$\\
    \hline
 \hline
         & \multicolumn{5}{c||}{\% of goals found at 20\% resources} \\
    \hline
    Robotics                     & $30.5\PM1.2$& $58.3\PM2.0$ & $83.2\PM1.8$ & $75.5\PM2.6$& $89.1\PM1.5$\\
    Students                     & $23.7\PM0.6$& $54.6\PM1.3$ & $65.4\PM1.7$ & $68.2\PM1.6$& $78.3\PM1.6$\\
    Mathematics                  & $28.3\PM0.8$& $60.3\PM2.9$ & $88.6\PM1.0$ & $69.7\PM2.5$& $92.0\PM1.1$\\
    Football                     & $42.1\PM1.7$& $71.5\PM0.8$ & $80.6\PM1.4$ & $71.6\PM1.1$& $91.3\PM1.4$\\
    Sports                       & $41.6\PM0.7$& $68.5\PM1.4$ & $78.5\PM1.6$ & $79.4\PM1.6$& $86.4\PM1.5$\\
    Laboratories                 & $18.1\PM0.8$& $80.4\PM3.1$ & $86.4\PM2.0$ & $86.0\PM1.3$& $92.1\PM1.4$\\
    Food                         & $48.5\PM1.7$& $92.7\PM2.2$ & $95.1\PM1.6$ & $89.1\PM1.8$& $95.0\PM1.7$\\
    Publications                 & $18.5\PM1.1$& $64.2\PM1.9$ & $87.2\PM2.1$ & $78.9\PM1.2$& $94.7\PM1.4$\\
    Projects                     & $32.0\PM1.4$& $80.5\PM2.1$ & $85.4\PM1.7$ & $89.8\PM2.5$& $95.2\PM1.7$\\
    News                         & $40.4\PM1.3$& $88.9\PM0.8$ & $93.8\PM1.2$ & $93.9\PM1.1$& $96.5\PM1.5$\\
    \hline
  \end{tabular}
  \caption{\small The performance of the two marginal-utility based methods
  with all the enhancements enabled. The numbers in parentheses are standard deviations.}\label{marginal-utility-final-table}
\end{center}
\end{table}

\begin{table}[!ht]
\small
\begin{center}
  \begin{tabular}{|c||c|c||c|c||c|c||}
 \hline
    Goal type  & \multicolumn{6}{c||}{\% of goals found} \\
    \cline{2-7}
    &\multicolumn{2}{c||}{Contract: } & \multicolumn{2}{c||}{Contract: } & \multicolumn{2}{c||}{Contract: } \\
    &\multicolumn{2}{c||}{5\% Resources} & \multicolumn{2}{c||}{10\% Resources} & \multicolumn{2}{c||}{20\% Resources} \\
    \cline{2-7}
                                & Anytime & Contract & Anytime & Contract & Anytime & Contract \\
    \hline
    Robotics                    &  $0.1\PM0.0$ &  $2.9\PM0.2$  & $18.1\PM0.7$ & $24.6\PM1.1$  & $58.3\PM2.0$ & $69.5\PM2.2$ \\
    Students                    &  $0.7\PM0.0$ &  $1.9\PM0.1$  & $17.0\PM1.9$ & $29.3\PM1.5$  & $54.6\PM1.3$ & $62.8\PM1.3$ \\
    Mathematics                 &  $1.8\PM0.3$ &  $8.4\PM0.8$  & $15.4\PM0.8$ & $24.1\PM0.9$  & $55.3\PM0.9$ & $64.0\PM1.1$ \\
    Football                    &  $4.2\PM0.2$ &  $12.1\PM0.5$ & $25.3\PM0.9$ & $29.9\PM1.0$  & $71.5\PM0.8$ & $75.6\PM0.8$ \\
    Sports                      &  $12.7\PM0.6$&  $20.4\PM1.1$ & $45.1\PM2.7$ & $60.4\PM1.1$  & $68.5\PM1.4$ & $77.7\PM1.3$ \\
    Laboratories                &  $10.5\PM1.0$&  $11.3\PM0.6$ & $33.4\PM0.7$ & $40.7\PM0.9$  & $80.4\PM3.1$ & $81.6\PM0.1$ \\
    Food                        &  $9.4\PM0.9$ &  $22.1\PM1.2$ & $30.3\PM1.7$ & $32.1\PM1.6$  & $92.7\PM2.2$ & $92.8\PM2.2$ \\
    Publications                &  $2.1\PM0.3$ &  $12.9\PM1.7$ & $35.2\PM2.6$ & $55.0\PM2.3$  & $64.2\PM1.9$ & $76.2\PM1.5$ \\
    Projects                    &  $0.9\PM0.0$ &  $21.5\PM1.6$ & $41.2\PM1.6$ & $48.6\PM0.9$  & $80.5\PM2.1$ & $86.5\PM1.6$ \\
    News                        &  $8.8\PM1.0$ &  $13.2\PM0.9$ & $22.6\PM0.8$ & $41.5\PM1.0$  & $88.9\PM0.7$ & $91.1\PM1.0$ \\
    \hline

    \hline
  \end{tabular}
  \caption{\small Contract vs. anytime performance when supplying a resource limit. The numbers in parentheses are standard deviations.}\label{limiting-resources}
\end{center}
\end{table}
\begin{table}[!ht]
\small
\begin{center}
  \begin{tabular}{|c||c|c||c|c||c|c||}
 \hline
    Goal type  & \multicolumn{6}{c||}{\% of resources consumed} \\
    \cline{2-7}
     &\multicolumn{2}{c||}{Contract: 5\% Goals} & \multicolumn{2}{c||}{Contract: 20\% Goals} & \multicolumn{2}{c||}{Contract: 50\% Goals} \\
    \cline{2-7}
                                & Anytime & Contract & Anytime & Contract & Anytime & Contract \\
    \hline
    Robotics                    &  $8.4\PM0.6$ &  $6.2\PM1.1$ & $12.1\PM0.7$& $9.8\PM0.5$  &  $17.2\PM1.2$ & $10.2\PM1.0$ \\
    Students                    &  $9.0\PM0.5$ &  $7.1\PM0.4$ & $11.3\PM0.5$& $8.4\PM0.3$  &  $18.7\PM0.6$ & $9.9\PM0.5$ \\
    Mathematics                 &  $5.4\PM0.4$ &  $3.9\PM0.2$ & $10.3\PM0.5$& $6.8\PM0.4$  &  $14.3\PM0.6$ & $9.9\PM0.4$ \\
    Football                    &  $5.2\PM0.2$ &  $2.1\PM0.2$ & $8.4\PM0.3$ & $5.2\PM0.2$  & $14.6\PM0.5$  & $9.9\PM0.7$ \\
    Sports                      &  $3.9\PM0.2$ &  $1.9\PM0.1$ & $7.1\PM0.6$ & $6.0\PM0.4$  & $12.2\PM0.6$  & $6.5\PM0.6$ \\
    Laboratories                &  $4.2\PM0.1$ &  $1.8\PM0.1$ & $7.4\PM0.4$ & $5.3\PM0.3$  & $14.1\PM0.6$  & $6.6\PM0.2$ \\
    Food                        &  $5.9\PM0.3$ &  $2.2\PM0.3$ & $8.8\PM0.5$ & $5.1\PM0.2$  & $14.3\PM0.5$  & $7.0\PM0.6$ \\
    Publications                &  $6.1\PM0.5$ &  $3.3\PM0.4$ & $7.7\PM1.1$ & $5.9\PM0.5$  & $18.8\PM0.7$  & $10.3\PM0.6$ \\
    Projects                    &  $4.0\PM0.3$ &  $1.7\PM0.1$ & $8.3\PM0.5$ & $6.1\PM0.3$  & $16.5\PM0.6$  & $9.0\PM0.5$ \\
    News                        &  $7.6\PM0.7$ &  $5.9\PM0.4$ & $9.8\PM0.2$ & $6.5\PM0.3$  & $16.0\PM0.4$ & $10.3\PM0.6$ \\
    \hline

    \hline
  \end{tabular}
  \caption{\small Contract vs. anytime performance when supplying goal requirements. The numbers in parentheses are standard deviations.}\label{limiting-goals}
\end{center}
\end{table}

\clearpage
\bibliography{davidov06a}
\bibliographystyle{theapa}
\end{document}